\documentclass{article}

\usepackage{tikz}
\usetikzlibrary{arrows.meta, positioning, calc}
\definecolor{lmSignal}{HTML}{5D7B9C}
\definecolor{lmProAlt}{HTML}{C0392B}
\definecolor{lmAltNoise}{HTML}{E8A4A0}
\definecolor{lmNoise}{HTML}{888888}
\definecolor{lmChaff}{HTML}{BBBBBB}
\definecolor{lmInk}{HTML}{1A2530}
\definecolor{lmPanelBlue}{HTML}{EBF1F8}
\definecolor{lmPanelYellow}{HTML}{FFF1D6}
\definecolor{lmPanelRose}{HTML}{FFEEEC}
\definecolor{lmPanelGreen}{HTML}{E8F2EB}
\definecolor{lmEdgeBlue}{HTML}{3A6090}
\definecolor{lmEdgeYellow}{HTML}{A07820}
\definecolor{lmEdgeRose}{HTML}{C0392B}
\definecolor{lmEdgeGreen}{HTML}{2E7A47}

 \usepackage[eandd, final]{neurips_2026}

\usepackage{amsmath} 
\usepackage[utf8]{inputenc} 
\usepackage[T1]{fontenc}    
\usepackage{hyperref}       
\usepackage{xurl}           
\usepackage{booktabs}       
\usepackage{amsfonts}       
\usepackage{nicefrac}       
\usepackage[expansion=false]{microtype} 
\usepackage{todonotes}
\usepackage{xcolor}         
\usepackage{graphicx}       
\usepackage{wrapfig}        
\usepackage{subcaption}     
\usepackage{listings}       
\usepackage{tcolorbox}      
\tcbuselibrary{listings,skins,breakable}
\newtcolorbox{simplegray}[1]{
  enhanced, breakable,
  colback=gray!6, colframe=gray!50,
  title=#1, fonttitle=\bfseries\small,
  boxrule=0.4pt, arc=1pt, left=4pt, right=4pt, top=3pt, bottom=3pt,
}
\lstdefinestyle{promptstyle}{
  basicstyle=\ttfamily\scriptsize,
  breaklines=true, breakatwhitespace=true,
  columns=fullflexible, keepspaces=true,
  frame=none, showstringspaces=false,
  upquote=true, escapechar=!,
}
\definecolor{albertcolor}{RGB}{0, 128, 0}       
\definecolor{andrewcolor}{RGB}{214, 90, 20}     
\definecolor{igorcolor}{RGB}{148, 0, 211}       
\definecolor{bishancolor}{RGB}{180, 0, 0}       
\definecolor{silcolor}{RGB}{0, 100, 180}        
\definecolor{tericolor}{RGB}{180, 120, 0}       
\definecolor{yongjaecolor}{RGB}{20, 160, 150}   

\title{\textsc{InvestigationWorlds}:\\ An Agentic Environment for Legal Investigation}

\author{
  Albert Yu Sun$^{1}$\thanks{Equal contribution.} \quad
  Andrew Benard$^{1}$\footnotemark[1] \quad
  Sil Hamilton$^{1,2}$\footnotemark[1] \quad
  Anna Teresita A. Marcelo$^{1}$ \\
  \bfseries Yong Jae Kim$^{1}$ \quad
  Carl-Leander Henneking$^{1}$ \quad
  Rundong Hu$^{1}$ \quad
  Yuhong Wang$^{1}$ \\
  \bfseries David Mimno$^{2}$ \quad
  Bishan Yang$^{1}$ \quad
  Igor Labutov$^{1}$ \\[6pt]
  {\normalfont $^{1}$Epiq AI Labs \quad $^{2}$Cornell University}
}

\begin{document}

\maketitle

\begin{abstract}
We introduce \textsc{InvestigationWorlds}, an agentic environment for legal investigation. 
We build on an underused artifact of U.S. civil litigation: the summary judgment motion.
This motion relies upon a record composed of real evidence exhibits, and results in a court-adopted hypothesis that is treated as ground truth for the purposes of deciding the motion. 
Each environment is built from a real U.S. Federal Court case retrieved from Public Access to Court Electronic Records (PACER) and augmented by an attorney-validated generation pipeline that synthesizes role-tagged documents around the original record.
The resulting corpus supports competing, internally coherent accounts of the case, requiring agents to distinguish hypotheses supported by the court-adopted findings from plausible alternatives.
Evaluating on 100 cases, we find that agents often endorse unsupported hypotheses despite retrieving relevant evidence.
\end{abstract}

\section{Introduction}

Many real-world reasoning tasks require synthesizing a coherent narrative from large collections of evidence that support multiple competing interpretations.
One example is legal discovery, the phase in a lawsuit when attorneys read across evidence to understand the narratives within it. 
By some estimates, discovery accounts for 90\% of litigation costs and time \citep{girard_espinosa_limiting_evasive_discovery, withers_computer_based_discovery}. 
We study this \textbf{investigation task}, where an agent must navigate a corpus of documents with minimal guidance and construct competing factual hypotheses of what happened.

No existing benchmarks measure this capability.
Current evaluations test legal question answering \citep{Zheng_Guha_Arifov_Zhang_Skreta_Manning_Henderson_Ho_2025}, precedent retrieval \citep{Wiratunga_Abeyratne_Jayawardena_Martin_Massie_Nkisi-Orji_Weerasinghe_Liret_Fleisch_2024}, hallucination in commercial research tools \citep{Magesh_Surani_Dahl_Suzgun_Manning_Ho_2024}, and bar-exam-style reasoning \citep{Guha_Nyarko}. 
Existing benchmarks typically assume a well-posed query, a single correct answer, or both. 
They rarely model situations in which the same body of evidence supports multiple coherent yet mutually incompatible conclusions.

We resolve the ground-truth problem by exploiting an underused artifact of U.S.\ civil litigation: the summary judgment motion. 
During such motions, both sides must lay out every material fact they intend to rely on, attach evidentiary exhibits supporting each fact, and either accept or dispute the opposing party's facts one by one. 
The judge issues a written opinion that adopts a specific set of factual findings, each traceable to the exhibits in the record. 
A decided summary judgment motion therefore provides a complete evidentiary corpus paired with a court-adopted hypothesis serving as judicially validated ground truth.

We use these summary judgment motions to build the core of \textsc{InvestigationWorlds}, an investigation environment derived from real federal cases obtained through Public Access to Court Electronic Records (PACER).
We surround the merits exhibits with synthetic documents to construct realistic, large corpora with controllable ambiguity and alternative hypotheses.
This setting exposes a failure mode not captured by existing evaluations: agents may retrieve relevant evidence yet still commit to incorrect alternative hypotheses.

\begin{figure}[t]
    \centering
    \includegraphics[width=\textwidth]{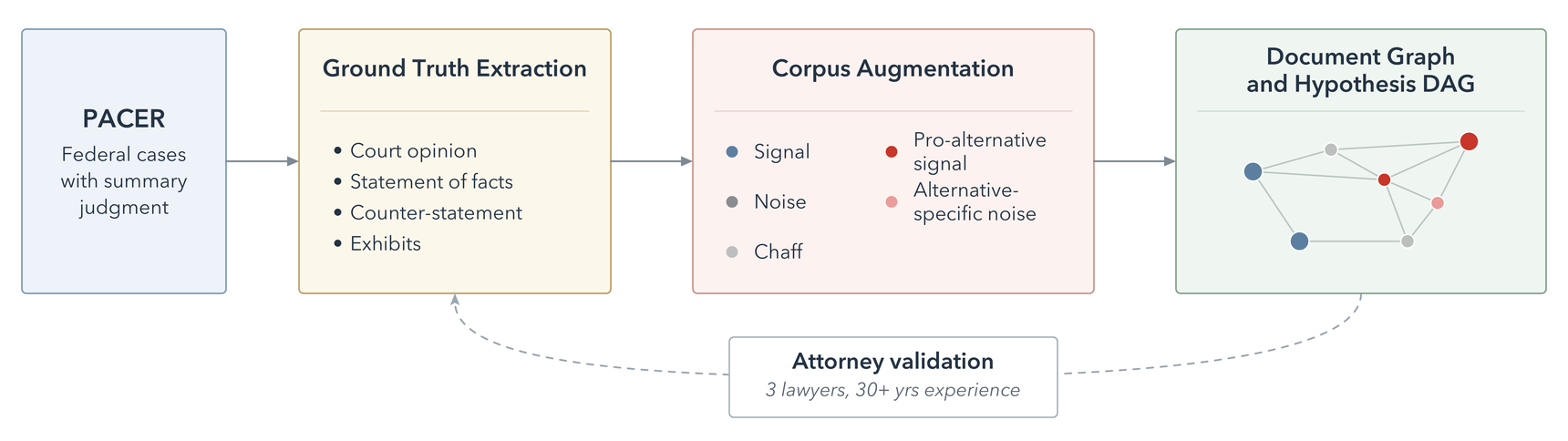}
    \caption{The \textsc{InvestigationWorlds} pipeline.
      Real federal cases with summary judgment motions are filtered from PACER.
      Statements of facts, supporting exhibits, and the Court's opinion are
      processed by an LLM pipeline that extracts entities, ground facts, and
      the two party hypotheses ($h_p$, $h_d$), and assembles them into a
      document graph and hypothesis Directed Acyclic Graph (DAG).
      A planner-and-judge agent then augments the document graph with synthetic
      exhibits tagged by evidentiary role (see \autoref{tab:document-roles}).}
    \label{fig:pipeline}
\end{figure}

We contribute the following:



\paragraph{Investigation task.}
\textsc{InvestigationWorlds} is the first agentic environment grounded in real court cases that requires agents to investigate competing factual interpretations, a central challenge in litigation.
Agents examine evidence that supports multiple plausible factual findings and predict the court-adopted disposition.
The court's opinion provides the ground truth for evaluation.

\paragraph{Dataset grounded in real cases.}
We release the full PACER document corpus for all 100 cases, along with anonymized versions of the generated synthetic documents.

\paragraph{Reproducible generation pipeline.}
We introduce an attorney-validated generation procedure (\autoref{fig:pipeline}) that transforms real case records into investigation corpora admitting multiple plausible factual findings, with each synthetic document tagged by its evidentiary role.
We release the generation code, all prompts, and the attorney-validated role taxonomy to support reproduction and extension to additional cases.
The dataset and pipeline code is available at \href{https://huggingface.co/datasets/epiq-ai-labs/InvestigationWorlds}{huggingface.co/datasets/epiq-ai-labs/InvestigationWorlds}.

\begin{figure}[t]
    \centering
    \includegraphics[width=\textwidth]{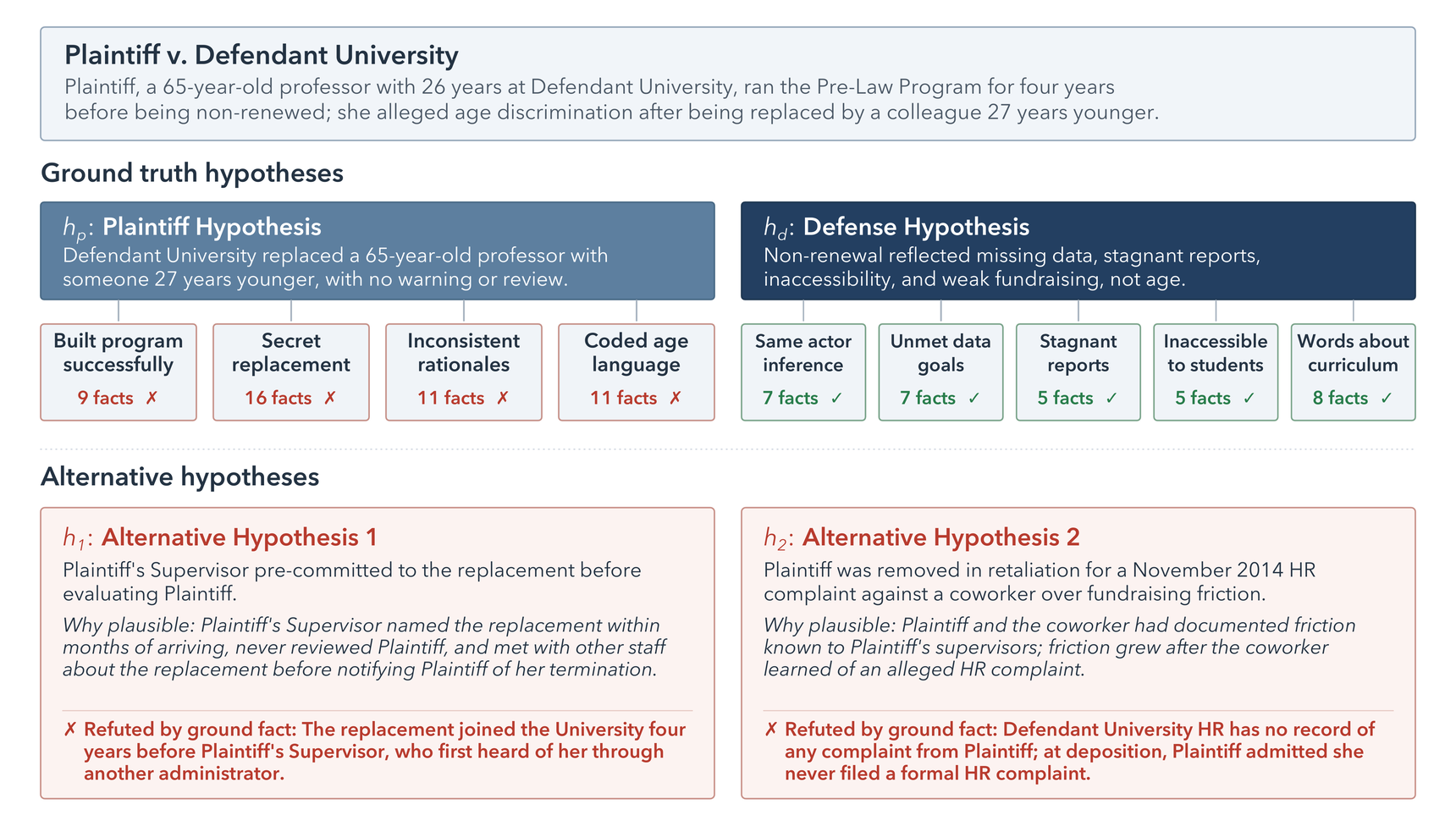}
    \caption{
    Example \textsc{InvestigationWorlds} hypothesis DAG from a representative employment-discrimination summary judgment motion. The banner summarizes the underlying dispute. \textbf{Ground truth hypotheses}: we show both party hypotheses, decomposed into sub-claims with the court-adopted/court-rejected ruling and supporting ground-fact counts. \textbf{Alternative hypotheses}: we show two synthetic alternative hypotheses such that each hypothesis is refutable by a ground fact embedded in the record.
    }
    \label{fig:worked-example}
\end{figure}

\subsection{The Investigation Task}
\label{sec:task}

We define an investigation environment as a triple $(\mathcal{D}, \mathcal{E}, \mathcal{H})$: a corpus of documents $\mathcal{D}$, a set of entities $\mathcal{E}$ that recur across them, and a hypothesis DAG $\mathcal{H}$ whose root is the dispositive legal question and whose leaves are ground facts supported or refuted by individual documents. 
The DAG $\mathcal{H}$ contains two party hypotheses and a set of \emph{alternative hypotheses}. The latter are plausible factual readings of the case, each partially backed by its own subset of $\mathcal{D}$, but ultimately refuted upon a complete reading of the record.

We use court-adopted hypothesis ($h^*$) for an account the judge adopted, court-rejected hypothesis ($h^\dagger$) for an account the judge rejected, party hypotheses for the pair ($\{h_p, h_d\}$), and alternative hypothesis  ($h_1, h_2$) for a synthetic distractor generated by our pipeline.
The agent has retrieval access to $\mathcal{D}$ and must commit to a position on each hypothesis in $\mathcal{H}$. 

This is an investigation task rather than simple QA because the agent must construct its account of the case before it sees the menu of hypotheses.
Success requires believing every positive party hypothesis and not believing any negative hypothesis, including synthetic alternatives constructed to distract from the ground truth. \autoref{fig:worked-example} shows an example.

\section{Constructing the Environment}
\label{sec:pipeline}

To create the investigation environments for \textsc{InvestigationWorlds}, we introduce the pipeline shown in \autoref{fig:pipeline} to identify cases, extract ground truth, and generate synthetic documents. Each stage is then validated by experienced attorneys. 
The full text of every prompt used in this pipeline is reproduced in Appendix~\ref{app:prompts}.

\subsection{Case Selection from PACER}

\begin{figure}[htbp]
\centering
\includegraphics[width=\linewidth]{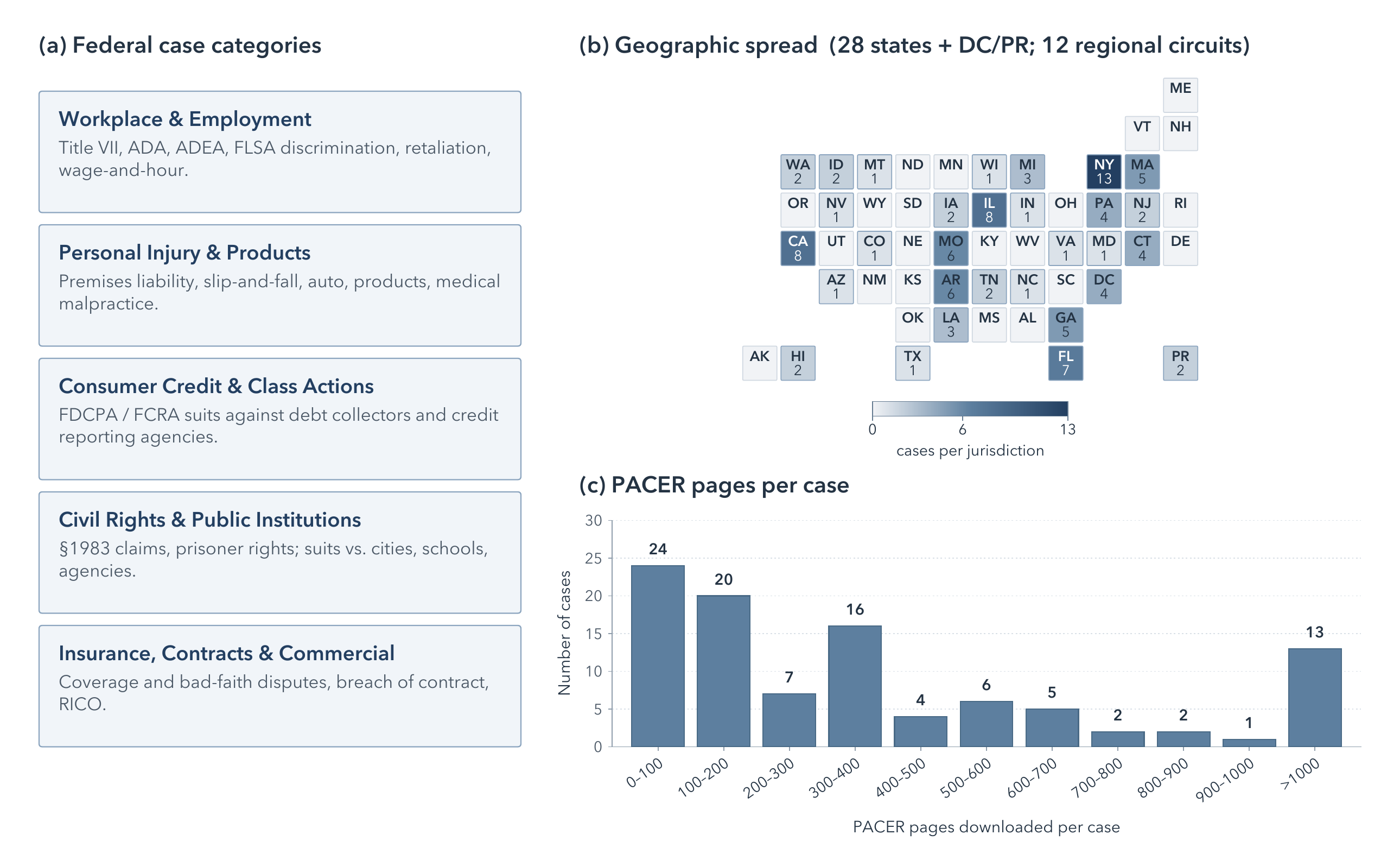}


\caption{\textbf{The \textsc{InvestigationWorlds} intended construction roster.} (a) 100 cases split evenly across five federal categories. (b) 28 states, the District of Columbia, and Puerto Rico across 41 federal judicial districts and all 12 regional circuits; shading encodes case count. (c) Downloaded PACER PDF pages per case (median 290, max 3{,}274; 42{,}283 total).}

\label{fig:dataset-overview}
\end{figure}

Our filter selects cases with fully decided summary judgment motions where the statements of facts, counter-statements of facts, evidentiary exhibits, and the court's opinion deciding the motion are available and unsealed.
The resulting cohort spans 100 cases from 28 states, the District of Columbia, and Puerto Rico across 41 federal judicial districts in all 12 regional circuits, balanced 20-per-category across 5 federal case types, with 42{,}283 pages in the archived downloaded PACER PDFs (\autoref{fig:dataset-overview}).
Because we retrieved PACER documents using the RECAP browser extension \citep{freelawprojectrecap, yu2011liberate}, every document we purchased has also been contributed to the Free Law Project's public RECAP Archive on CourtListener \citep{courtlistener}.

For the documents we make accessible to the agent, we restrict the exhibit list to \emph{merits exhibits}, which contain the evidence that bears on the summary judgment motion.
Specifically, we discard filings that are primarily procedural in nature or those which would provide the ground truth answers to the agent (e.g., expert opinions).

\subsection{Ground Truth Extraction}

A language model reads the opinion and extracts the factual findings the court actually made, the legal questions it had to decide, and the specific textual passages that support each finding.
In parallel, the pipeline parses the statements of facts and the opposing party's counter-statement, which together enumerate the facts the parties either agreed on or contested.

We use an LLM (Claude Sonnet 4.6) to read across the remaining corpus to extract entities such as the named people, organizations, instruments, and artifacts that appear in the case.
We disambiguate them so that references to the same real-world actor across different documents collapse to a single identity.
Then, we use the LLM to annotate each exhibit with its document type, its relevant date, and the entities it mentions.

\subsection{Augmenting the Corpus}

A case's merits exhibits are far smaller than the corpora agents face in real legal investigations.
To convert a real summary judgment record into an investigation environment, we augment the merits exhibits with documents that make the corpus larger and introduce controllable alternative hypotheses. 
A language model (Claude Sonnet 4.6 by default; ablated in \autoref{sec:generator-robustness}) generates two alternative hypotheses $(h_1, h_2)$ per case.
Each alternative hypothesis is paired with the reasoning that makes it superficially plausible and the facts that refute it.

This language model reads the full case context, including the story, the canonical facts, the entities, the alternative hypotheses, the text of every merits exhibit in the record, and it drafts a slate of document specifications. 
The model is instructed to plan documents in coherent series.
For example, a series could be a purchase order, the matching invoice, and the corresponding wire-transfer record that all share participants and exist close together in time.

Each specification designates a role, a target fact or hypothesis, and a rendering format (pdf, docx, xlsx, pptx, or plain text) chosen to mirror the formats of real discovery productions. 
The five roles are described in \autoref{tab:document-roles}.
Sub-agents then render each document so that it reads as a plausible artifact from the case world.
A judge model (Claude Haiku 4.5) reviews each finished document to ensure it meets the specification. Documents that fail are rejected and regenerated up to three times.

\begin{table}[htbp]
\centering
\small
\caption{The five roles assigned to synthetic documents.}
\label{tab:document-roles}
\begin{tabular}{@{}lp{0.72\textwidth}@{}}
\toprule
\textbf{Name} & \textbf{Description} \\
\midrule
Signal & Embeds one or more ground facts in a realistic discovery artifact, anchoring at least one node in the case's fact-DAG amid ambient material consistent with the merits exhibits. \\
\addlinespace
Noise & Shares parties, contracts, and date windows with signal documents but embeds no ground facts. Tests whether retrieval distinguishes on-topic material from important evidence. \\
\addlinespace
Chaff & Belongs to the same business world as the case but shares none of its parties, dates, or dispute topics. Adds the administrative clutter real productions contain (facilities notices, vendor invoices, HR memos). \\
\addlinespace
Pro-alternative signal & Embeds a fact that tilts the agent toward an alternative hypothesis while remaining individually compatible with every ground fact. The adversarial counterpart to signal. \\
\addlinespace
Alternative-specific noise & Consistent with the parties and timeline an alternative hypothesis implies, but embeds none of its facts. Provides topical cover for that alternative. \\
\bottomrule
\end{tabular}
\end{table}

\subsection{Attorney Validation}
\label{sec:attorney-validation}

Before generation, a litigation attorney pre-screened the case set to confirm that each case had requirements for a factually disputed summary judgment record.
Three practicing attorneys with 30+ years of combined experience then audited the pipeline outputs.
They completed approximately 100 hours of expert review.

After generation, all 100 cases received a hypothesis-fidelity audit by one of the attorneys. 
They labeled whether the two extracted party hypotheses faithfully capture the court-adopted and court-rejected hypotheses and whether the two alternative hypotheses are realistic readings of the same record.
This pass ensures that for every case, the task is grounded in the judicial record and the alternative hypotheses we ask agents to discriminate between are plausible.

In addition to the cohort-wide audit, 55 of the 100 cases received a stage-by-stage review covering the following artifacts that our pipeline produces: party hypotheses, alternative hypotheses, signal/noise documents, chaff, and document metadata. 
For the actual document validation, the attorneys reviewed the general document plan for each hypothesis (an overview plus a brief description of each document).
Then, they validated a subset of documents of each role for each case in detail.
Within these 55 cases, an agreement subset of 10 cases (drawn proportionally across the five federal categories) was independently reviewed by all three attorneys.

For the 55 case subset, pass rates exceeded 94\% for the party hypotheses (and their related documents), alternative hypotheses (and their related documents), and chaff documents. 
This is achieved at a 81\%+ agreement score, which was computed on the 10-case agreement subset.
Per-stage pass rates are reported in Table~\ref{tab:validation-summary} in Appendix~\ref{app:annotator}.

To verify that our approach yields valid solutions and that each court opinion is attainable from the merits exhibits, an attorney investigated 10 randomly selected cases across the 5 categories described in \autoref{fig:dataset-overview}, none of which they had audited during the original validation. With the court's opinion and adopted label withheld, the attorney received only the merits evidence, reading 459 pages total over 12 hours. Their factual conclusions agreed with the court-adopted outcome in all 10 cases.

\section{Experimental Setup}
\label{sec:experimental-setup}

\subsection{Tasks and Protocol}

\paragraph{Two-turn investigation.}
In \emph{Turn 1}, the agent receives the case type and the document corpus. It is instructed to investigate the corpus, identify the most likely account of what happened, identify alternative scenarios the evidence could support, cite documents, and produce an investigation summary.
Tools are enabled.

In \emph{Turn 2}, the agent receives only its own Turn-1 summary and a shuffled list of hypotheses labeled \texttt{H1}, \texttt{H2}, \ldots 
\textbf{Tools are disabled}, forcing the agent to commit based on the investigation it conducted in Turn 1 rather than re-searching the corpus to fit the offered hypotheses.
For each hypothesis, the agent returns one of three verdicts: \emph{believe}, \emph{do not believe}, or \emph{do not know}.
The two-turn separation is what makes the task an investigation rather than a simple QA or retrieval task. 
The agent must build its account before seeing the menu of options.
Prompts for both turns are in Appendix~\ref{app:prompts}.

\paragraph{Hypothesis set and scoring.}
The hypothesis set for case $c$ is $\mathcal{H}_c = \{h_p, h_d, h_1, h_2\}$, composed of two party hypotheses ($h_p$, $h_d$) extracted from the plaintiff's and defense's summary judgment briefs and two alternative hypotheses ($h_1$, $h_2$) generated by our pipeline.
Let $Y_c \subseteq \{h_p, h_d\}$ denote the set of positive party hypotheses for case $c$; either or both party hypotheses may belong to $Y_c$.
The negative hypotheses are $\mathcal{H}_c \setminus Y_c$.
The ground-truth verdict is \emph{believe} for every hypothesis in $Y_c$ and \emph{do not believe} for every hypothesis in $\mathcal{H}_c \setminus Y_c$.

We define our success metric, \textbf{case-solved}, to score 1 if the model believes every hypothesis in $Y_c$ and does not believe any hypothesis in $\mathcal{H}_c \setminus Y_c$.
A ``do not know'' verdict on a negative hypothesis counts as not believing it, but a ``do not know'' verdict on a positive hypothesis does not satisfy the success criterion.
This metric tests whether a language model identifies the full set of positive party hypotheses from the case documents while avoiding belief in negative hypotheses.

We also compute per-class accuracy (believe-class accuracy and do-not-believe-class accuracy), whose true labels are ``believe'' and ``do not believe'', respectively.
For these class-specific accuracies, ``do not know'' is incorrect in both classes.

For a case with $k=|Y_c|$ positive hypotheses, uniformly random independent verdicts achieve case-solved probability $(1/3)^k(2/3)^{4-k}$.
A policy that believes exactly one uniformly selected hypothesis and rejects the rest succeeds with probability $1/4$ when $k=1$ and zero when $k=2$.
We average each baseline over the cases evaluated for each model.

\subsection{Models, Harnesses, and Resources}

We evaluate six frontier agents: Claude Sonnet 4.6 and Claude Haiku 4.5 via the Claude Agent SDK; GPT 5.5 and GPT 5.4 mini via the OpenAI Agents SDK; and DeepSeek V4 Flash and Qwen 3.6 35B via a vLLM-backed mini-swe-agent harness exposing OpenAI-compatible chat completions.
Both harnesses expose the same four file-system operations (read, glob, grep, shell).
Each agent is allotted a fixed budget per task for turn 1.
After the tool budget is exhausted, further tool requests are denied and the agent is instructed to produce its final output.
Compute environment, concurrency, and per-call telemetry are detailed in Appendix~\ref{app:implementation}.

\section{Results}
\label{sec:results}


\subsection{Investigation}

\begin{figure}[htbp]
    \centering
    \includegraphics[width=\textwidth]{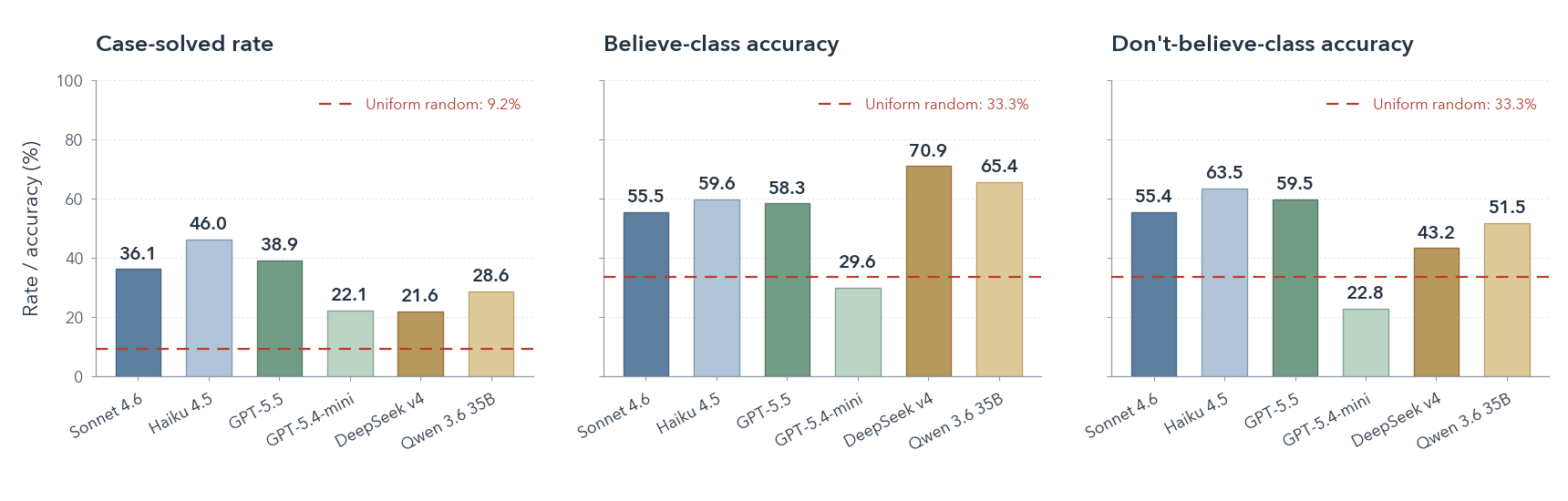}
    \caption{Performance on the 100-case cohort: case-solved rate (left), believe-class accuracy (center), and do-not-believe-class accuracy (right). Dashed lines show the uniform-random verdict baseline: 9.2\% (rounded) for case-solved rate and 33.3\% for each class accuracy. This policy selects uniformly among the three verdict options (\textit{believe}, \textit{do not believe}, \textit{do not know}).}
    \label{fig:investigation}
\end{figure}


\paragraph{The best-performing model solves fewer than half of cases.}
\autoref{fig:investigation} reports case-solved rate and class-specific
accuracy on the 100-case cohort. Haiku 4.5 achieves the highest
case-solved rate (46.0\%), followed by GPT-5.5 (38.9\%) and
Sonnet 4.6 (36.1\%).
All six models exceed the uniform-random verdict baseline (9.2\%).
Five models exceed the stronger baseline that believes one
hypothesis at random and rejects the rest
(21.4\%--21.6\%, depending on the evaluated cases);
DeepSeek V4 Flash matches this baseline.

Class-specific accuracy helps distinguish performance on positive
hypotheses from performance on negative hypotheses. The always-reject
policy achieves 100\% accuracy on negative hypotheses but 0\% on
positive hypotheses. Five of the six models achieve
43.2\%--59.6\% accuracy on their worse-performing class, exceeding
the maximum worst-class accuracy of 33.3\% among the evaluated
naive policies. GPT-5.4-mini is the exception, with a worst-class
accuracy of 22.8\%.

\paragraph{Commitment behavior contextualizes model rankings.}
On the 87 cases with complete outputs from both models, Haiku 4.5
commits on 80.7\% of verdicts, compared with 73.3\% for Sonnet 4.6.
Haiku is also more accurate among its committed verdicts
(77.2\% versus 76.1\%).
Case-solved performance additionally depends on avoiding belief
in negative hypotheses. Haiku and Sonnet believe all positive
hypotheses in 49 and 46 cases, respectively, but also believe a
negative hypothesis in 9 and 15 of those cases.
Consequently, Haiku solves 40/87 cases (46.0\%), compared with
31/87 (35.6\%) for Sonnet.

\subsection{Counterfactual Ablations over Document Roles}


To assess the effects of corpus composition, we run a five-level
cumulative ablation on 50 cases. Starting from a \emph{merits-only}
baseline containing the original merits exhibits and no synthetic
documents, we add 20 pro-alternative documents, approximately 57 noise
documents, and chaff at 50\% and 100\% of the maximum load
(approximately 250 and 500 chaff documents in total, respectively).

\begin{figure}[t]
    \centering
    \includegraphics[width=\textwidth]{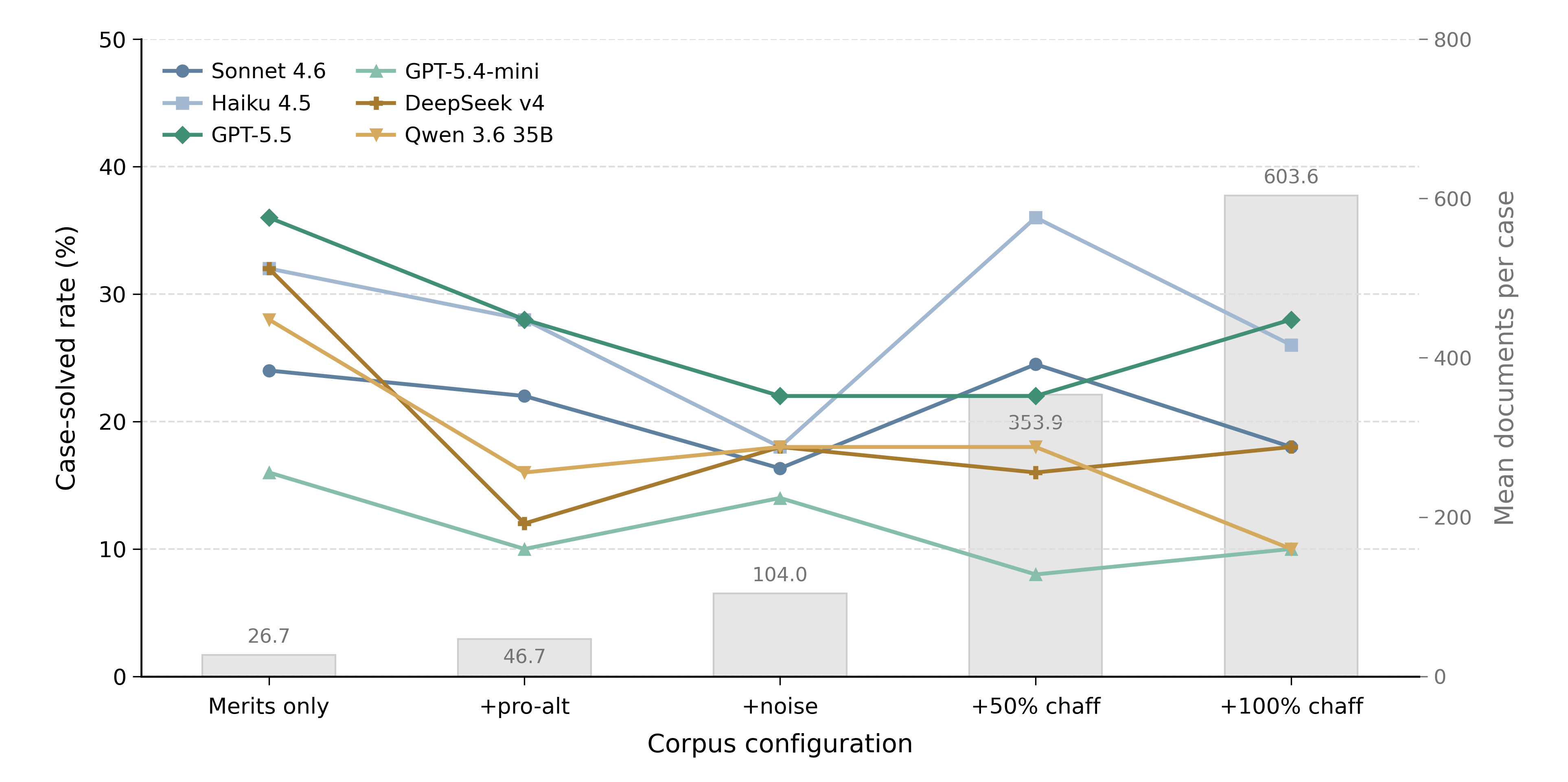}
    \caption{Corpus size ablation for a subset of 50 cases. Case-solved rates decrease for all six models as pro-alternative documents are added; additional noise and chaff produce no consistent further decline. Sonnet's noise and 50\%-chaff conditions each contain 49 completed
runs after one run exceeded the two-hour limit.}
    \label{fig:ablations}
\end{figure}

Adding pro-alternative documents reduces case-solved rates for all
six models (\autoref{fig:ablations}). The declines range from 2 to
20 percentage points; the paired 95\% bootstrap confidence interval
excludes zero only for DeepSeek V4 Flash ($-20$ points;
95\% CI: $[-36, -4]$). This pattern is consistent with
alternative hypotheses being easier to reject when the corpus
contains no documents supporting them.

\begin{figure}[t]
    \centering
    \includegraphics[width=\textwidth]{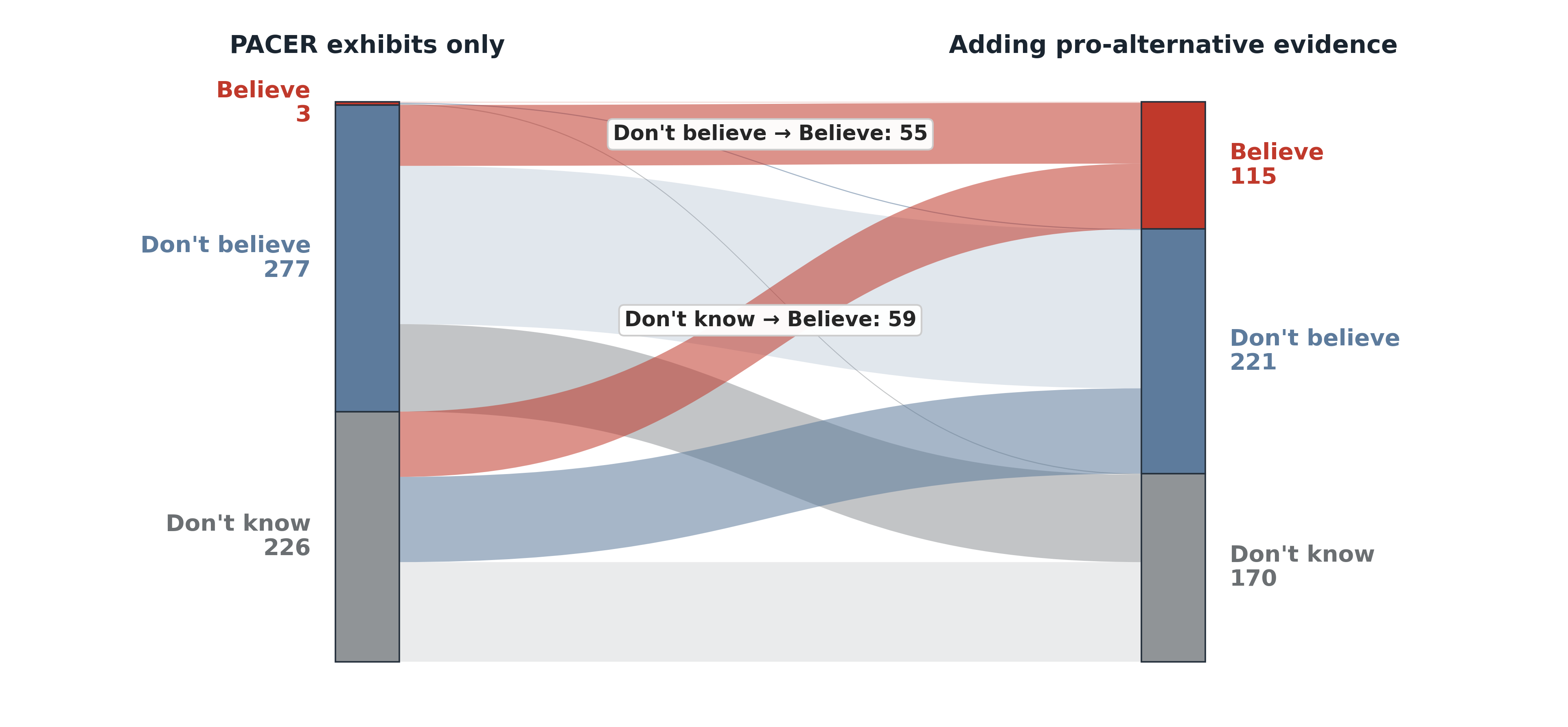}
\caption{Verdict shifts for synthetic alternative hypotheses when
pro-alternative evidence is added to the merits-only corpus. From six
models, 50 cases, and two alternatives per case, we retain 506 paired
verdicts with parsed answers in both conditions, excluding 94 pairs with
missing/unparsed output in either condition. Belief in these
reference-incorrect alternatives increases from 3/506 (0.6\%) to
115/506 (22.7\%): 55 answers change from \textit{do not believe},
59 from \textit{do not know}, and one remains \textit{believe}.
Ribbons are colored by destination.}

    \label{fig:sankey}
\end{figure}

Using the same 50-case ablation, \autoref{fig:sankey} examines how verdicts on alternative hypotheses change when pro-alternative evidence is added to a merits-only corpus.
Across six models and two alternatives per case, we analyze 506 paired verdicts with parsed answers in both conditions.
Adding 20 pro-alternative documents increases belief in an alternative hypothesis from 3/506 (0.6\%) to 115/506 (22.7\%).
Of the 114 verdicts that switch to \textit{believe}, 55 previously expressed \textit{do not believe} and 59 expressed \textit{do not know}; one additional verdict remains \textit{believe}.
These shifts suggest that pro-alternative evidence can induce commitment to alternative hypotheses that agents previously rejected or remained uncertain about.

\subsection{Investigative Strategies}

\begin{figure}[t]
    \centering
    \includegraphics[width=0.7\textwidth]{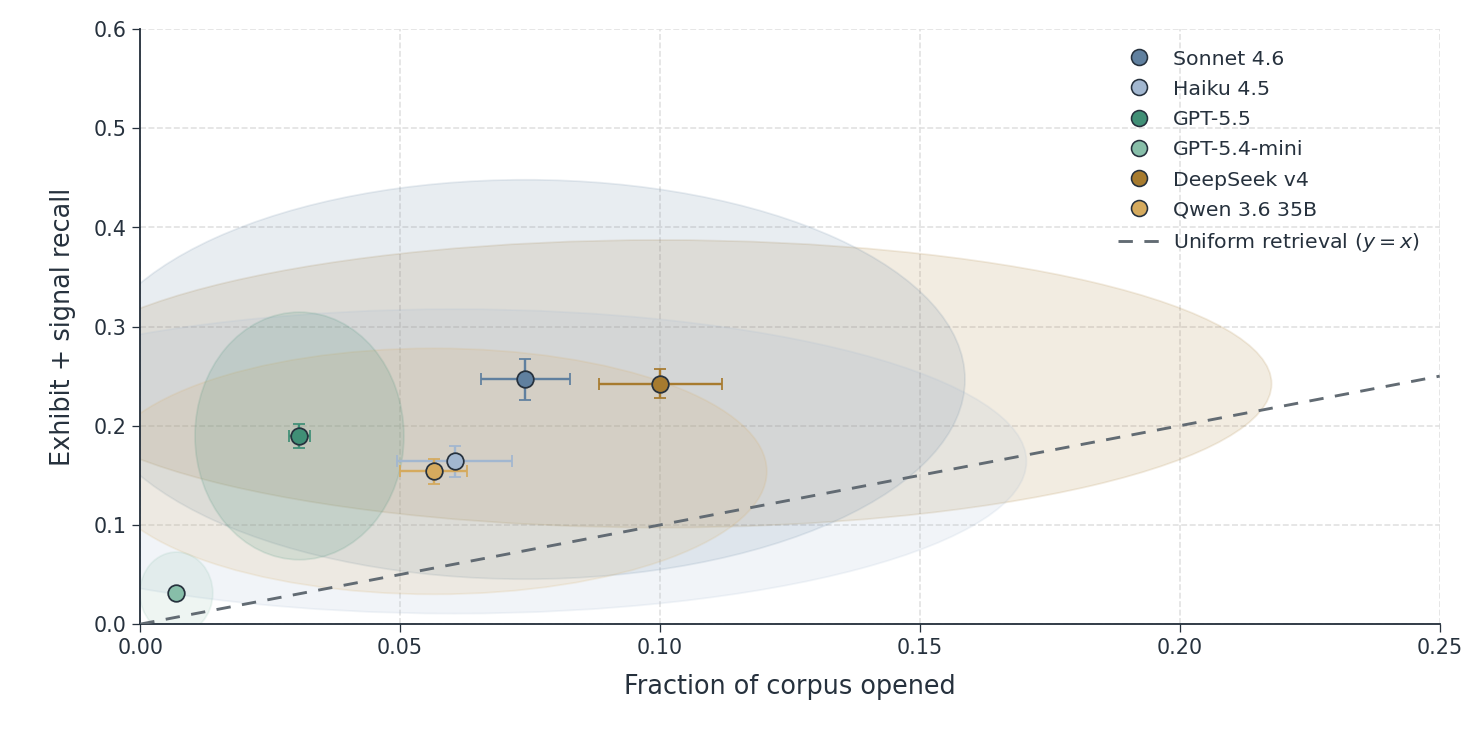}
    \caption{Per-model mean Turn-1 trajectory. $x$-axis: fraction of corpus opened; $y$-axis: fraction of merits exhibits and $h^*$-supporting signal documents opened. Bars are $\pm$1\, Standard Error of the Mean (SEM); ellipses span $\pm$1\,$\sigma$.}
    \label{fig:trajectories}
\end{figure}

We characterize investigative strategy along two axes: \emph{document coverage} (fraction of corpus the agent opens) and \emph{signal-document recall} (fraction of merits evidence and $h^*$-supporting signal documents opened).
In \autoref{fig:trajectories}, every model sits well above the uniform-retrieval reference, consistent with selective entity-driven retrieval rather than exhaustive sweeps: agents open 0.7--10\% of the corpus on average yet recover signal at substantially higher rates.
Two models anchor the high-effort corner: Sonnet 4.6 (7.4\% coverage, 24.7\% signal recall) and DeepSeek V4 Flash (10.0\% coverage, 24.2\% signal recall). DeepSeek actually reads more documents on average, with Sonnet hitting signal at a slightly higher rate per document opened.
Haiku, Qwen-3.6-35B, and GPT-5.5 form a middle cluster at 3--6\% coverage and 15--19\% signal recall, while GPT-5.4 mini sits alone near the origin (0.7\% coverage, 3.1\% signal recall).
Despite a large spread in coverage and recall across models, none of these strategy differences predict accuracy, reinforcing that retrieval volume is not the bottleneck.



\subsection{Robustness to Generator Choice}
\label{sec:generator-robustness}

A natural concern is that \textsc{InvestigationWorlds} measures
alignment between evaluator and generator rather than
investigative ability.
To examine this concern, we rebuilt 30 held-out cases end-to-end
using OpenAI models in place of Anthropic models throughout the
pipeline (ground-truth extraction, alternative hypothesis
generation, signal/noise/chaff document rendering, and the judge).
We then re-evaluated six models on both versions of each case:
Sonnet 4.6, Haiku 4.5, GPT-5.5, GPT-5.4-mini, DeepSeek v4,
and Qwen 3.6 35B.

Using case-solved rate, we find that \textbf{evaluator rankings
are positively correlated across generator conditions
(Spearman $\rho=0.77$)}.
One pair of evaluators reverses its strict ordering:
DeepSeek v4 overtakes Qwen 3.6 35B on OpenAI-generated corpora.
Haiku 4.5 leads on Anthropic-generated corpora at 40.0\% and
ties GPT-5.5 at 40.0\% on OpenAI-generated corpora.
DeepSeek v4 shows the largest observed change in case-solved
rate, from 20.0\% on Anthropic-generated corpora to 36.7\% on
OpenAI-generated corpora (paired difference: +16.7 percentage
points; 95\% CI: $[-3.3, 36.7]$).

These results show variation in point-estimate rankings across
the two tested generation pipelines. Paired 95\% confidence
intervals for differences in case-solved rate include zero for
all six evaluators.
\autoref{fig:generator-robustness} in \autoref{app:robustness}
shows the comparison.

\section{Related Work}

\paragraph{Agentic environments and reasoning benchmarks.}
A recent wave of agentic environments evaluates models on procedurally generated tasks with verifiable rewards \citep{reasoninggym, kor, gem, textarena}, where ground truth is computable by an algorithmic verifier and corpora are synthesized end-to-end.
\textsc{InvestigationWorlds} differs because the underlying matter is sourced from a real matters.
GAIA \citep{mialon2023gaia} evaluates multi-step tool use and web browsing on grounded questions, and BrowseComp \citep{wei2025browsecomp} stresses persistent multi-hop retrieval for deep-research agents; both reward locating a single objective answer rather than committing to one of several coherent factual readings constructed adversarially against the agent.
The closest work is the agentic-coding family \citep{yao2024tau,jimenez2023swe}, which similarly evaluates trajectories, but in a domain where the reward is a passing test suite rather than a contested factual reading.

\paragraph{Multi-document reasoning.}
There exist prior work on building benchmarks surrounding realistic multi-hop scenarios. 
CorporateBench \citep{hamilton2026corporatebench} evaluates information extraction and question answering over large synthetic corporate document collections generated from temporally evolving knowledge bases.
Multi-hop QA benchmarks introduced the distractor-document evaluation pattern. 
HotpotQA \citep{yang2018hotpotqa} pairs gold supporting paragraphs with TF-IDF distractors, and MuSiQue \citep{trivedi2022musique} composes single-hop questions to control for shortcut reasoning. 
MuSR \citep{sprague2024musr} extends this line to long-form narratives generated by a neurosymbolic pipeline with adversarial distractors.
CONFLICTS \citep{cattan2025dragged} extends evaluation to conflicting retrieved sources, using expert annotations of knowledge-conflict types to assess how search-augmented models handle such discrepancies.
\textsc{InvestigationWorlds} differs because it is anchored in a real federal case with judicially validated ground truth and distractors are typed by \emph{evidentiary role} against a hypothesis DAG rather than by retrieval-similarity to a query. 

\paragraph{Legal AI benchmarks.}
Existing legal benchmarks primarily evaluate retrieval-augmented generation (RAG) for question answering or assess general legal knowledge, but do not test narrative reconstruction from heterogeneous evidence.
RAG-based legal benchmarks span bar-exam and housing-law QA \citep{Zheng_Guha_Arifov_Zhang_Skreta_Manning_Henderson_Ho_2025}, retrieval precision and recall on legal documents \citep{pipitone2024legalbenchrag}, precedent retrieval for case-based reasoning \citep{Wiratunga_Abeyratne_Jayawardena_Martin_Massie_Nkisi-Orji_Weerasinghe_Liret_Fleisch_2024}, hallucination in commercial legal tools \citep{Magesh_Surani_Dahl_Suzgun_Manning_Ho_2024}, and contract review against junior lawyers \citep{Martin_Whitehouse_Yiu_Catterson_Perera_2024}.
Other benchmarks evaluate general legal reasoning, including concept memorization \citep{Fei_Shen_Zhu_Zhou_Han_Zhang_Chen_Shen_Ge_2023}, general legal tasks \citep{Guha_Nyarko}, Chinese legal tasks \citep{Yu_Lin_Ni_Cheng_Sha_2025}, and hallucination \citep{Dahl_Magesh_Suzgun_Ho_2024}.
LEGIT \citep{lee2026legit} evaluates legal reasoning traces using hierarchical issue-tree rubrics derived from court judgments, measuring issue coverage and reasoning correctness.
Harvey LAB \citep{harveylab2026} evaluates legal agents on matter-based tasks requiring reviewable work products from supplied documents, assessed using expert-written rubrics.
\textsc{InvestigationWorlds} targets legal investigation, a capability that existing benchmarks have not directly measured.

\section{Discussion and future work}

\paragraph{Scope of jurisdiction and case type.}
Our environments are drawn from U.S.\ federal civil cases that reached a fully briefed summary judgment motion. 
This is a focused but tractable slice of legal practice, since the summary judgment posture provides exactly the structured ground truth that makes rigorous evaluation possible. 
The methodology generalizes to other domains with similar evidentiary structures: state-court records, administrative adjudications, and arbitration awards. 
We see \textsc{InvestigationWorlds} as a first instantiation of a broader template for building investigation environments wherever institutional decisions create reliable ground truth.

\paragraph{Synthetic documents are attorney-validated, not court-tested.}
Our pipeline produces documents that experienced litigators judge plausible and role-consistent, a standard sufficient for the multi-hypothesis reasoning task that motivates the benchmark. 
Forensic-style realism (metadata patterns, threading conventions, native-format artifacts) can be layered onto the existing pipeline incrementally, and the role-tagging framework will continue to apply as document fidelity improves.

\paragraph{Judicial rulings as ground truth.}
Our labels are procedural rather than absolute: a summary judgment opinion does not resolve contested facts, but identifies the factual account the record supports under the Rule 56 standard.
We therefore scope ground truth to the account the court adopted for purposes of deciding the motion, and the benchmark measures agreement with that reading of the record.
Two checks support this choice.
First, an attorney given only the merits exhibits reached the court-adopted outcome in all 10 sampled cases (\autoref{sec:attorney-validation}), indicating that the label is recoverable from the evidence agents see.
Second, we audited the appellate history of all 100 cases as of 27 July 2026: 22 had appellate activity, and in none did an appellate decision alter the factual account we use.
Absence of reversal is not proof of correctness, since most rulings are never appealed, and labels should be revisited if later proceedings disturb them.

\subsection{Release and Ethical Considerations}
\label{sec:release}

The 100 source cases are U.S. federal court filings obtained via PACER and are public records; copies are also mirrored on CourtListener's RECAP archive. We re-host the full document corpus on Hugging Face so that the benchmark is self-contained and reproducible without per-case PACER retrieval.

The synthetic documents generated by our pipeline interpolate real entity names drawn from these filings (attorneys, parties, organizations) into fabricated artifacts such as invoices, emails, and internal memos.
Although every named entity already appears in the corresponding public record, juxtaposing real names with fabricated content creates reputational and privacy risk.
We therefore release the synthetic corpus only in pseudonymized form, alongside the full PACER corpus, the generation pipeline, prompts, and role taxonomy, and withhold versions containing original names. A reader with API access to the listed models can regenerate a synthetic corpus from the released pipeline and PACER corpus.

\newpage
\begingroup
\apptocmd{\thebibliography}{\raggedright}{}{}
\bibliography{bibliography}
\bibliographystyle{plainnat}
\endgroup

\appendix

\section{Implementation Details}
\label{app:implementation}

\paragraph{Agent harnesses.}
Claude runs use the Claude Agent SDK with the model aliases \texttt{opus}, \texttt{sonnet}, and \texttt{haiku}, exposing the SDK's native \texttt{Read}, \texttt{Bash}, \texttt{Glob}, and \texttt{Grep} tools.
The SDK is configured with a maximum buffer size of 10\,MB.
OpenAI runs use the OpenAI Agents SDK with the model identifiers as listed in the main text, exposing function tools \texttt{read\_file}, \texttt{glob\_files}, \texttt{grep\_files}, and \texttt{bash}.
The OpenAI SDK is configured with \texttt{truncation="auto"}.

\paragraph{Resource budgets.}
The Claude factual-QA harness uses a maximum of 15 agent turns and a 30-tool-call budget.
The OpenAI factual-QA harness uses a maximum of 50 agent turns since it hit MaxTurnsExceeded errors on 75\% of the cases when max turns was 25. It also has a 30-tool-call budget.
The investigation harness for both providers uses 25 agent turns in Turn 1 with a 30-tool-call budget; once the tool budget is exhausted, further tool requests are denied and the model is instructed to produce its final summary.
Tools are disabled in Turn 2 for all runs.
No wall-clock timeout is imposed beyond the SDK and tool-call limits, except for the two-hour limit in the expanded 50-case ablation.

\paragraph{Compute environment.}
Frontier-API models (Claude Sonnet 4.6, Claude Haiku 4.5, GPT 5.5, GPT 5.4 mini) are accessed through the Anthropic and OpenAI APIs from a commodity Linux orchestration host.
Open-weight models (DeepSeek V4 flash, Qwen 3.6 35B) are served by a local vLLM deployment on NVIDIA GPUs and exposed at OpenAI-compatible endpoints.
No specialized storage is required beyond the case corpus on local disk.

\paragraph{Concurrency and scheduling.}
The evaluation runner enforces per-model semaphores (Anthropic and OpenAI each capped at three concurrent requests) inside a twelve-worker executor pool, with exponential backoff on rate-limit responses.
Construction-pipeline jobs are dispatched through a tmux pool with at most three concurrent case pipelines.

\paragraph{Per-call telemetry and total compute.}
Every LLM call in both the construction pipeline and the evaluation harnesses is logged to a file containing input, output, and cache token counts, wall-clock duration, and dollar cost; aggregate totals can be reproduced by summing these files across the released cohort.
The full project consumed approximately \$3{,}300 of API spend, with per-case investigation costs ranging from \$2.06 (Sonnet 4.6) to \$0.07 (GPT 5.4 mini); open-weight runs incur only local GPU time.

\section{Annotator Results}
\label{app:annotator}

For attorney validation, we receive the following scores shown in \autoref{tab:validation-summary}. Besides document metadata extraction, all stages critical to the investigation task exceed 94\% pass and 81\% pairwise agreement, with pooled agreement of 78.2\% across all stages.

\begin{table}[htbp]
\centering
\small
\caption{Attorney validation. Coverage and judgment rates aggregate the deep-review audit across all 55 cases; document-metadata coverage (45 cases) trails because it relies on a separately scheduled pass. Pairwise observed agreement is computed on the 10-case agreement subset, where all three attorneys independently rated the same items.}
\label{tab:validation-summary}
\begin{tabular}{@{}lrrrrrr@{}}
\toprule
& & & \multicolumn{3}{c}{Judgment distribution (\%)} & \\
\cmidrule(lr){4-6}
Stage & Cases & Items & Pass & Borderline & Fail & \%\,agree \\
\midrule
Party hypotheses              & 55 & 150     & 98.0 & 0.0  & 2.0  & 93.3  \\
Alternative hypotheses        & 55 & 308     & 96.4 & 0.0  & 3.6  & 92.1  \\
Chaff                         & 55 & 225     & 94.7 & 3.6  & 1.8  & 81.1  \\
Document metadata             & 45 & 1{,}326 & 73.7 & 15.8 & 10.5 & 72.6  \\
\midrule
All stages (pooled) & 55 & 2{,}009 & 81.3 & 10.9 & 7.8 & 78.2 \\
\bottomrule
\end{tabular}
\end{table}

\section{Generator Robustness}
\label{app:robustness}
\begin{figure}[t]
    \centering
    \includegraphics[width=0.85\textwidth]{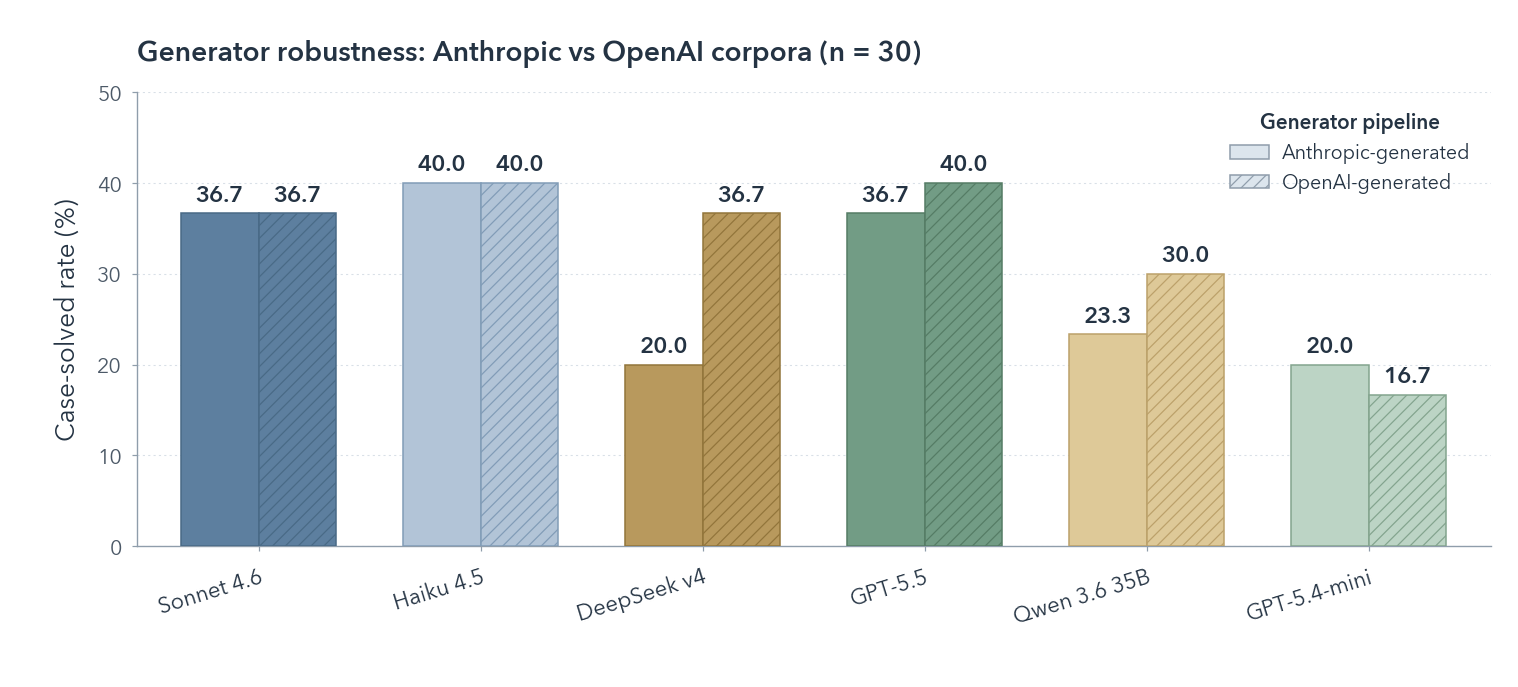}
    \caption{Case-solved rates for six evaluators on $n=30$ held-out
cases with environments generated using Anthropic vs.\ OpenAI
pipelines, including the open-weight evaluators DeepSeek v4 and
Qwen 3.6 35B. Rankings are positively correlated across generator
conditions (Spearman $\rho=0.77$, using average ranks for ties),
with one strict reversal: DeepSeek v4 overtakes Qwen 3.6 35B.}
    \label{fig:generator-robustness}
\end{figure}

As shown in \autoref{fig:generator-robustness}, the relative ordering of DeepSeek v4 and Qwen 3.6 35B reverses across generator conditions.

\begingroup
\raggedbottom
\section{Prompt Library}
\label{app:prompts}

This appendix reproduces the prompts used in the \textsc{InvestigationWorlds} pipeline, organized by stage. 
Each prompt is a Jinja2 template; tokens such as \texttt{\{\{ opinion\_text \}\}} or \texttt{\{\% include \ldots\ \%\}} are filled at runtime with case-specific content.

\subsection{Ground-Truth Extraction}
\label{app:prompts:extract}

The prompts in this section convert a real federal summary judgment record (the judge's opinion, the parties' Statement of Undisputed Facts and response, and the briefs) into the canonical ground-truth artifacts on which every downstream stage depends: a fact table, party hypothesis DAGs, and the judge's ruling annotation.

\paragraph{Opinion Pass 1.}
First-pass opinion analysis. Loaded by \path{extract_utils/opinion_analysis.py} to extract the argument map, merits-relevant facts the court relies on, and atomic factual question stubs from the judge's written opinion.

\begin{simplegray}{Opinion Pass 1 Prompt (\texttt{prompts/extract/opinion\_pass1.j2})}
\begin{lstlisting}[escapechar=!]
You are a legal analyst. Decompose this judge's opinion into an argument map, extract the merits-relevant facts the court relies on, and stub every question the downstream artifact will contain.

**Output a single JSON object** with this shape:

```
{
  "case": "...",
  "citation": "...",
  "facts": [
    {
      "fact_id": "op_MARKETCAP_LAUNCH",
      "source_type": "opinion",
      "text": "<one-sentence factual statement>",
      "cited_exhibit": "Doc. 57-11" | null,
      "highlight_text": "<10-30 word VERBATIM quote from opinion>",
      "opinion_page": <int>,
      "merits": true
    }
  ],
  "arguments": [
    {
      "id": "arg_1",
      "label": "<legal question in plain language>",
      "argument_type": "factual_analysis" | "procedural_defense",
      "opinion_page": <int>,
      "branches": [
        {
          "id": "arg_1a",
          "label": "<short descriptive name>",
          "opinion_page": <int>,
          "sources": [
            {
              "id": "src_57_11",
              "doc": "Doc. 57-11",
              "description": "<date, type, title>",
              "question_stubs": [
                {
                  "id": "MARKETCAP_LAUNCH",
                  "fact_ids": ["op_MARKETCAP_LAUNCH"],
                  "question": "<natural-language question>",
                  "answer_sketch": "<one-line answer>"
                }
              ]
            }
          ]
        }
      ]
    }
  ]
}
```

**Extract facts first.**

- A fact is an atomic claim about the world the court treats as established. One sentence per fact; split compound findings.
- **Only include merits facts** -- facts that bear on the substantive legal question the court is resolving. Exclude procedural facts (filing dates, motion history), standard-of-review recitations, and pure background. Test: would this fact change if the court's ruling came out the other way? If no, exclude it.
- **Only include underlying facts, not litigation-stage expert characterizations.** Facts must describe events that occurred before the lawsuit -- who did what, when, where, and what was communicated. Exclude post-hoc expert opinions offered during litigation that characterize or classify the underlying conduct (e.g., "Dr. X opined that the treatment fell below the standard of care"). These are not part of the factual world this exercise targets.
- Each fact must include `highlight_text`: a 10-30 word **verbatim** quote from the opinion containing the claim. Copy exactly -- punctuation, capitalization, everything. Do not paraphrase. This is the anchor downstream stages rely on.
- `fact_id` is `op_` + a SCREAMING_SNAKE_CASE descriptive suffix, unique across the whole artifact.
- `opinion_page` comes from the `[PAGE N]` markers in the opinion text.
- `cited_exhibit` is the evidence the court cites for the fact, or null if none.

**Then decompose the opinion into arguments and branches.**

- Each top-level legal question the court resolves is an `argument`. Use the court's own section headers where available.
- **Classify each argument's `argument_type`:**
  - `"factual_analysis"` -- the argument analyzes the underlying conduct using evidentiary exhibits (declarations, depositions, documents, exhibits). This includes the core substantive questions the case turns on (e.g., "Did the defendant sell securities?", "Was the agency's search adequate?").
  - `"procedural_defense"` -- the argument addresses a threshold legal question citing only case law and statutes, not evidentiary exhibits. Examples: fair notice defenses, standing challenges, statute of limitations, jurisdictional questions, collateral estoppel. The test: does resolving this argument require examining what the parties actually did (factual_analysis), or only what the law requires (procedural_defense)?
- Within each argument, identify the court's analytical sub-moves -- these are `branches`. A sub-move is a distinct line of reasoning (e.g., "what the defendant said publicly", "how the defendant's business model worked", "why the defendant's counter-defense fails").
- Within each branch, list every evidentiary document the court cites. Skip citations used only for legal authority (case law, statutes, treatises).
- If the same document supports multiple branches, split it into distinct source entries with disambiguating ids (e.g., `src_61_3_structure`, `src_61_3_uses`).
- If an argument is resolved purely on precedent with no evidentiary sub-analysis, set `branches: []` and add a top-level `note` field explaining why.

**Then stub every question.**

- For each source, write one `question_stub` per discrete factual claim the opinion draws from it. Each stub references the fact(s) it asks about by `fact_ids`. Most stubs reference exactly one fact.
- Question ids are SCREAMING_SNAKE_CASE, unique across the whole artifact, and descriptive. Use shared prefixes for related questions (`COO_EMAIL_SUBJECT`, `COO_EMAIL_OPPORTUNITY`, `COO_EMAIL_TIMEFRAME`).
- `answer_sketch` is a one-line answer -- just enough for the downstream pass to recognize what the question is asking. No polish, no rubric.
- **Do not write synthesis or cross-source questions.** Each question must ask about a single fact. Do not write questions that relate or connect facts from multiple sources.

**CRITICAL: Write atomic factual questions, not legal, compound, or synthesis questions.**

The downstream benchmark tests whether a system can extract information from evidentiary documents (contracts, letters, declarations, agreements). Every question stub must satisfy ALL of these requirements:

1. **Factual, not legal.** The question must ask about FACTS found in documents -- not about the court's legal conclusions, statutory interpretations, or how facts apply to law. Never frame a question as "What does the court say about..." or "What did the court conclude about..." or "What does the opinion say about..." -- these all ask about the court's analysis, not about the underlying documents. Instead ask "What does [the exhibit/declaration/agreement] say about..." Test: "Could someone answer this by reading the cited exhibit alone, or does it require reading the judge's opinion?" If the latter, reframe.
2. **Atomic, not compound.** Each question must ask exactly ONE thing. If a question contains "and" joining two distinct inquiries, or asks about multiple facts, split it into separate stubs. Test: "Does this question have exactly one answer, or does it have two answers to two different sub-questions?" If the latter, split.
3. **Does not reveal too much about where to find the answer.** Strip specifics from questions -- do not name the exhibit, clause, email, or document type when possible. Do not include so much detail that the question telegraphs its own answer. The reader should have to find the relevant information, not just confirm what the question already describes. Ask about the factual finding, not about "specific language" in the document.
4. **No synthesis or cross-source questions.** Each question must ask about a single fact that the court found in its opinion -- not about a conclusion drawn by connecting multiple facts. If a question requires relating two or more facts to reach a conclusion (even an obvious one), it is a synthesis question and must be excluded. Red-flag phrases: "taken together," "in light of," "considering both," "what pattern," "what do these collectively show." Even if two facts appear next to each other in the opinion with an implied causal relationship, do not ask about that relationship unless the court explicitly stated the conclusion.
5. **No contention framing.** Do not ask about what a party "contends," "alleges," or "argues." Ask about the underlying fact itself. A party's contention is not a fact; the fact is the thing they are contending about.
6. **Only underlying facts, not litigation-stage expert characterizations.** Include facts about what actually happened (who did what, when, where, what was communicated). Exclude post-hoc expert opinions offered during litigation that characterize or classify the care -- these are not part of the factual world this exercise targets. Test: "Did this fact exist before the lawsuit was filed?" If no, exclude it.

Here are examples of bad stubs and their corrected versions:

BAD (compound -- asks two separate things joined by "and"):
{
  "id": "AUTH_LETTER_TIMING_AND_CONTENT",
  "fact_ids": ["op_AUTH_LETTER_TIMING", "op_AUTH_LETTER_RIGHTS"],
  "question": "In what temporal sequence were the Authorization Letter and the Artist Agreement executed, and what does the Authorization Letter say about rights allocation?",
  "answer_sketch": "Auth Letter came first; it says Leibovitz holds all rights."
}
GOOD (split into two atomic stubs, each asking one thing):
{
  "id": "AUTH_LETTER_EXECUTION_DATE",
  "fact_ids": ["op_AUTH_LETTER_TIMING"],
  "question": "When was the Authorization Letter executed relative to the Artist Agreement?",
  "answer_sketch": "The Authorization Letter predates the Artist Agreement."
}
{
  "id": "AUTH_LETTER_RIGHTS_LANGUAGE",
  "fact_ids": ["op_AUTH_LETTER_RIGHTS"],
  "question": "What does the Authorization Letter say about who holds rights?",
  "answer_sketch": "Leibovitz represents she holds all rights, title, and interest."
}

BAD (legal -- asks how facts compare to a legal standard):
{
  "id": "ARTIST_AGREEMENT_RIGHTS_VS_COPYRIGHT_BUNDLE",
  "fact_ids": ["op_ARTIST_AGREEMENT_NOT_AMONG_SIX_RIGHTS"],
  "question": "How do the rights in the Artist Agreement's grant clause compare to the standard bundle of copyright ownership rights?",
  "answer_sketch": "License/market/promote are not among the six s 106 rights."
}
GOOD (factual -- asks only what the agreement says):
{
  "id": "ARTIST_AGREEMENT_GRANTED_RIGHTS",
  "fact_ids": ["op_ARTIST_AGREEMENT_NOT_AMONG_SIX_RIGHTS"],
  "question": "What specific rights does the Artist Agreement's grant clause convey to Great Bowery Inc.?",
  "answer_sketch": "The exclusive worldwide right to license, market, and promote the Licensed Images."
}

BAD (asks what the court says -- the answer is the court's legal analysis, not a document fact):
{
  "id": "KERATOCONUS_OBJECTIVE_PRONG",
  "fact_ids": ["op_KERATOCONUS_SERIOUS"],
  "question": "What does the court say about whether Plaintiff's keratoconus satisfies the objective prong of the serious medical condition test?",
  "answer_sketch": "The court concludes keratoconus satisfies the objective prong."
}
GOOD (asks about the underlying fact):
{
  "id": "KERATOCONUS_DIAGNOSIS",
  "fact_ids": ["op_KERATOCONUS_SERIOUS"],
  "question": "What medical condition did the Plaintiff suffer from at the time of the events at issue?",
  "answer_sketch": "Keratoconus, with risk of progressive vision loss without treatment."
}

BAD (too much context -- names the exhibit, author, topic, and exact thing discussed):
{
  "id": "REDDY_SEPT2021_NO_UW_SECURE_FACILITY",
  "fact_ids": ["op_REDDY_NO_UW_FACILITY"],
  "question": "According to Dr. Reddy's September 2021 report, was corneal crosslinking available at UW's secure facility?",
  "answer_sketch": "No, UW did not offer crosslinking at its secure facility."
}
GOOD (stripped -- asks the same thing without giving away where to look):
{
  "id": "REDDY_SEPT2021_NO_UW_SECURE_FACILITY",
  "fact_ids": ["op_REDDY_NO_UW_FACILITY"],
  "question": "Was corneal crosslinking available at UW?",
  "answer_sketch": "No, UW did not offer crosslinking at its secure facility."
}

BAD (synthesis -- draws a causal connection the court only implied):
{
  "id": "SOUTH_MARCH2021_EQUIPMENT_LIMITATION_REFERRAL_LINK",
  "fact_ids": ["op_SOUTH_NO_EQUIPMENT", "op_SOUTH_REFERRAL"],
  "question": "How did Dr. South's lack of diagnostic equipment relate to her decision to submit a referral?",
  "answer_sketch": "She lacked the equipment, so she referred the patient out."
}
GOOD (atomic -- asks about each fact separately):
{
  "id": "SOUTH_NO_DIAGNOSTIC_EQUIPMENT",
  "fact_ids": ["op_SOUTH_NO_EQUIPMENT"],
  "question": "What equipment was Dr. South's facility lacking?",
  "answer_sketch": "Equipment needed to conduct topography testing."
}

BAD (contention framing -- asks what a party contends, not the underlying fact):
{
  "id": "KERATOCONUS_PLAINTIFF_CONTENTION",
  "fact_ids": ["op_KERATOCONUS_SERIOUS"],
  "question": "What does the Plaintiff contend is a serious medical need?",
  "answer_sketch": "Plaintiff contends keratoconus is a serious medical need."
}
GOOD (asks about the fact itself):
{
  "id": "KERATOCONUS_DIAGNOSIS",
  "fact_ids": ["op_KERATOCONUS_SERIOUS"],
  "question": "What medical condition did the Plaintiff suffer from at the time of the events at issue?",
  "answer_sketch": "Keratoconus."
}

BAD (asks about "specific language" rather than the factual finding):
{
  "id": "KURT_MARCH2022_CROSSLINKING_CHARACTERIZATION",
  "fact_ids": ["op_KURT_CROSSLINKING_BENEFIT"],
  "question": "What specific language did Dr. Kurt use to describe the potential benefit of crosslinking?",
  "answer_sketch": "Kurt described crosslinking as potentially preventing the need for transplant."
}
GOOD (asks about factual findings):
{
  "id": "KURT_MARCH2022_CROSSLINKING_CHARACTERIZATION",
  "fact_ids": ["op_KURT_CROSSLINKING_BENEFIT"],
  "question": "According to Dr. Kurt, what benefits (if any) could crosslinking have provided to the Plaintiff?",
  "answer_sketch": "Crosslinking could have prevented the need for a corneal transplant."
}

Return only the JSON object. No prose, no markdown fences.

OPINION:
{{ opinion_text }}
\end{lstlisting}
\end{simplegray}

\paragraph{Opinion Pass 2.}
Second-pass opinion analysis. Loaded by \path{extract_utils/opinion_analysis.py} after Pass 1 to enforce a stricter "lawsuit-events-only" filter and synthesize the structured opinion artifact (facts, arguments, and question stubs) used by the rest of the pipeline.

\begin{simplegray}{Opinion Pass 2 Prompt (\texttt{prompts/extract/opinion\_pass2.j2})}
\begin{lstlisting}[escapechar=!]
You are a legal analyst. Your goal is to review the Court's opinion and extract as many individual facts as possible that relate to the events that are the subject of the lawsuit and break them down into individual factual units. The secondary goal is to exclude any judicial findings that may be facts, but which do not relate to the events that are the subject of the lawsuit, but which instead relate to the lawsuit itself. Examples of the kinds of fact that relate to the lawsuit itself and not the subject of the lawsuit include but are not limited to: procedural facts such as when documents were filed, served or demanded; facts about what the law says or interpretations of what the law means; facts that require the application of the law to the facts; expert opinions about the facts; or conclusions drawn by synthesizing multiple facts together.

Decompose this judge's opinion into an argument map, extract the merits-relevant facts the court relies on, and stub every question the downstream artifact will contain.

**Output a single JSON object** with this shape:

```
{
  "case": "...",
  "citation": "...",
  "facts": [
    {
      "fact_id": "op_MARKETCAP_LAUNCH",
      "source_type": "opinion",
      "text": "<one-sentence factual statement>",
      "cited_exhibit": "Doc. 57-11" | null,
      "highlight_text": "<10-30 word VERBATIM quote from opinion>",
      "opinion_page": <int>,
      "merits": true
    }
  ],
  "arguments": [
    {
      "id": "arg_1",
      "label": "<legal question in plain language>",
      "argument_type": "factual_analysis" | "procedural_defense",
      "opinion_page": <int>,
      "branches": [
        {
          "id": "arg_1a",
          "label": "<short descriptive name>",
          "opinion_page": <int>,
          "sources": [
            {
              "id": "src_57_11",
              "doc": "Doc. 57-11",
              "description": "<date, type, title>",
              "question_stubs": [
                {
                  "id": "MARKETCAP_LAUNCH",
                  "fact_ids": ["op_MARKETCAP_LAUNCH"],
                  "question": "<natural-language question>",
                  "answer_sketch": "<one-line answer>"
                }
              ]
            }
          ]
        }
      ]
    }
  ]
}
```

**Extract facts first.**
The Court's opinion will be divided into sections. One section of the opinion will contain a recitation of the facts that are the subject of the lawsuit. Only extract facts from this section. Extract each individual fact that you can identify in the Court's opinion broken down into as informationally compact unit as possible (maximize the number of facts). Do not include procedural facts such as when documents were filed, served or demanded; facts about what the law says or interpretations of what the law means; facts that require the application of the law to the facts; expert opinions about the facts; or conclusions drawn by synthesizing multiple facts together.

- Each fact must include `highlight_text`: a 10-30 word **verbatim** quote from the opinion containing the claim. Copy exactly -- punctuation, capitalization, everything. Do not paraphrase. This is the anchor downstream stages rely on.
- `fact_id` is `op_` + a SCREAMING_SNAKE_CASE descriptive suffix, unique across the whole artifact.
- `opinion_page` comes from the `[PAGE N]` markers in the opinion text.
**Then stub every question.**
- write one `question_stub` per discrete factual claim the opinion draws from it. Each stub references the fact(s) it asks about by `fact_ids`. Most stubs reference exactly one fact.
- Question ids are SCREAMING_SNAKE_CASE, unique across the whole artifact, and descriptive. Use shared prefixes for related questions (`COO_EMAIL_SUBJECT`, `COO_EMAIL_OPPORTUNITY`, `COO_EMAIL_TIMEFRAME`). - `answer_sketch` is a one-line answer -- just enough for the downstream pass to recognize what the question is asking. No polish, no rubric.
- **Do not write synthesis or cross-source questions.** Each question must ask about a single fact. Do not write questions that relate or connect facts from multiple sources.
**CRITICAL: Write atomic factual questions, not legal, compound, or synthesis questions.**
- A fact is an atomic claim about the world the court treats as established. One sentence per fact; split compound findings.
- **Only include merits facts** -- facts that bear on the substantive legal question the court is resolving. Exclude procedural facts (filing dates, motion history), standard-of-review recitations, and pure background. Test: would this fact change if the court's ruling came out the other way? If no, exclude it.
- **Only include underlying facts, not litigation-stage expert characterizations.** Facts must describe events that occurred before the lawsuit -- who did what, when, where, and what was communicated. Exclude post-hoc expert opinions offered during litigation that characterize or classify the underlying conduct (e.g., "Dr. X opined that the treatment fell below the standard of care"). These are not part of the factual world this exercise targets.
- Each fact must include `highlight_text`: a 10-30 word **verbatim** quote from the opinion containing the claim. Copy exactly -- punctuation, capitalization, everything. Do not paraphrase. This is the anchor downstream stages rely on.
- `fact_id` is `op_` + a SCREAMING_SNAKE_CASE descriptive suffix, unique across the whole artifact.
- `opinion_page` comes from the `[PAGE N]` markers in the opinion text.
- `cited_exhibit` is the evidence the court cites for the fact, or null if none.

**Then decompose the opinion into arguments and branches.**

- Each top-level legal question the court resolves is an `argument`. Use the court's own section headers where available.
- **Classify each argument's `argument_type`:**
  - `"factual_analysis"` -- the argument analyzes the underlying conduct using evidentiary exhibits (declarations, depositions, documents, exhibits). This includes the core substantive questions the case turns on (e.g., "Did the defendant sell securities?", "Was the agency's search adequate?").
  - `"procedural_defense"` -- the argument addresses a threshold legal question citing only case law and statutes, not evidentiary exhibits. Examples: fair notice defenses, standing challenges, statute of limitations, jurisdictional questions, collateral estoppel. The test: does resolving this argument require examining what the parties actually did (factual_analysis), or only what the law requires (procedural_defense)?
- Within each argument, identify the court's analytical sub-moves -- these are `branches`. A sub-move is a distinct line of reasoning (e.g., "what the defendant said publicly", "how the defendant's business model worked", "why the defendant's counter-defense fails").
- Within each branch, list every evidentiary document the court cites. Skip citations used only for legal authority (case law, statutes, treatises).
- If the same document supports multiple branches, split it into distinct source entries with disambiguating ids (e.g., `src_61_3_structure`, `src_61_3_uses`).
- If an argument is resolved purely on precedent with no evidentiary sub-analysis, set `branches: []` and add a top-level `note` field explaining why.

**Then stub every question.**

- For each source, write one `question_stub` per discrete factual claim the opinion draws from it. Each stub references the fact(s) it asks about by `fact_ids`. Most stubs reference exactly one fact.
- Question ids are SCREAMING_SNAKE_CASE, unique across the whole artifact, and descriptive. Use shared prefixes for related questions (`COO_EMAIL_SUBJECT`, `COO_EMAIL_OPPORTUNITY`, `COO_EMAIL_TIMEFRAME`).
- `answer_sketch` is a one-line answer -- just enough for the downstream pass to recognize what the question is asking. No polish, no rubric.
- **Do not write synthesis or cross-source questions.** Each question must ask about a single fact. Do not write questions that relate or connect facts from multiple sources.

**CRITICAL: Write atomic factual questions, not legal, compound, or synthesis questions.**

The downstream benchmark tests whether a system can extract information from evidentiary documents (contracts, letters, declarations, agreements). Every question stub must satisfy ALL of these requirements:

1. **Factual, not legal.** The question must ask about FACTS found in documents -- not about the court's legal conclusions, statutory interpretations, or how facts apply to law. Never frame a question as "What does the court say about..." or "What did the court conclude about..." or "What does the opinion say about..." -- these all ask about the court's analysis, not about the underlying documents. Instead ask "What does [the exhibit/declaration/agreement] say about..." Test: "Could someone answer this by reading the cited exhibit alone, or does it require reading the judge's opinion?" If the latter, reframe.
2. **Atomic, not compound.** Each question must ask exactly ONE thing. If a question contains "and" joining two distinct inquiries, or asks about multiple facts, split it into separate stubs. Test: "Does this question have exactly one answer, or does it have two answers to two different sub-questions?" If the latter, split.
3. **Does not reveal too much about where to find the answer.** Strip specifics from questions -- do not name the exhibit, clause, email, or document type when possible. Do not include so much detail that the question telegraphs its own answer. The reader should have to find the relevant information, not just confirm what the question already describes. Ask about the factual finding, not about "specific language" in the document.
4. **No synthesis or cross-source questions.** Each question must ask about a single fact that the court found in its opinion -- not about a conclusion drawn by connecting multiple facts. If a question requires relating two or more facts to reach a conclusion (even an obvious one), it is a synthesis question and must be excluded. Red-flag phrases: "taken together," "in light of," "considering both," "what pattern," "what do these collectively show." Even if two facts appear next to each other in the opinion with an implied causal relationship, do not ask about that relationship unless the court explicitly stated the conclusion.
5. **No contention framing.** Do not ask about what a party "contends," "alleges," or "argues." Ask about the underlying fact itself. A party's contention is not a fact; the fact is the thing they are contending about.
6. **Only underlying facts, not litigation-stage expert characterizations.** Include facts about what actually happened (who did what, when, where, what was communicated). Exclude post-hoc expert opinions offered during litigation that characterize or classify the care -- these are not part of the factual world this exercise targets. Test: "Did this fact exist before the lawsuit was filed?" If no, exclude it.

[Bad/good question-stub examples are identical to Opinion Pass 1 above and are not duplicated here.]

Return only the JSON object. No prose, no markdown fences.

OPINION:
{{ opinion_text }}
\end{lstlisting}
\end{simplegray}

\paragraph{SOUF / RSOUF Fact Extraction.}
Loaded by \texttt{extract\_utils/souf\_facts.py}. Parses the movant's Statement of Undisputed Facts and the opposing party's response, classifying each numbered fact as undisputed or disputed and capturing exhibit citations.

\begin{simplegray}{SOUF/RSOUF Fact-Extraction Prompt (\texttt{prompts/extract/fact\_extraction\_souf\_rsouf.j2})}
\begin{lstlisting}[escapechar=!]
You are a legal analyst. You are given two documents:
1. A Statement of Undisputed Facts (SOUF) filed by the {{ souf_filer }}
2. The opposing party's Response to the Statement of Undisputed Facts

The SOUF lists numbered facts that the movant claims are undisputed. The Response admits or denies each fact.

For each fact in the SOUF, extract:
- fact_id: The fact number as stated in the SOUF (e.g., "1", "2", "3")
- text: The factual statement as written in the SOUF
- response_type: "undisputed" if the responding party admits/does not contest it, "disputed" if they deny or qualify it
- disputed_reason: If disputed, briefly state why (e.g., "Defendant denies the amount was $50,000" or "Plaintiff disputes the timeline"). Use "" if undisputed.
- cited_exhibit: Any exhibit referenced in support of this fact (e.g., "Ex. A", "Doc. 57-11") or null
- souf_page: Page number in the SOUF where this fact appears (from [PAGE X] markers)
- highlight_text: A short verbatim quote (10-30 words) from the SOUF containing this fact

Return a JSON array. No markdown, just the JSON array:
[
  {"fact_id": "1", "text": "...", "response_type": "undisputed", "disputed_reason": "", "cited_exhibit": "Ex. A", "souf_page": 2, "highlight_text": "..."},
  {"fact_id": "2", "text": "...", "response_type": "disputed", "disputed_reason": "Defendant denies...", "cited_exhibit": null, "souf_page": 2, "highlight_text": "..."},
  ...
]

STATEMENT OF UNDISPUTED FACTS (filed by {{ souf_filer }}):
{{ souf_text }}

RESPONSE TO STATEMENT OF UNDISPUTED FACTS (filed by {{ response_filer }}):
{{ response_text }}
\end{lstlisting}
\end{simplegray}

\paragraph{Ground Facts.}
Loaded by \texttt{extract\_utils/hypothesis\_packet.py}. Reconciles the opinion, SOFs, RSOFs, and briefs into a single canonical, atomic, citation-bearing fact table -- the factual propositions referenced by every hypothesis (real and alternative).

\begin{simplegray}{Ground-Facts Prompt (\texttt{prompts/extract/ground\_facts.j2})}
\begin{lstlisting}[escapechar=!]
You are extracting the canonical fact table for a summary judgment case. These facts are the load-bearing factual propositions that the case turns on -- they will be referenced by the plaintiff's hypothesis, the defense's hypothesis, and (in alt generation) by alternative hypotheses.

A **ground fact** is a factual proposition (who did what when where) that is sourced from at least one of:
- The judge's factual recitation in the opinion
- A Statement of Facts (SOF) or Response to SOF (RSOF) by either party
- The MSJ memorandum, opposition brief, or reply brief

Facts must be:
- **Atomic** -- one event/state/relationship per fact. Split compound facts.
- **Factual, not legal** -- describe observable events, actions, and states. Do NOT embed legal standards, statutory terms, or conclusions-of-law as descriptors, even when the opinion uses that language. Rewrite to the underlying observable predicate.
- **Citation-bearing** -- each fact references at least one source document with page/paragraph if available.
- **Deduplicated** -- if two sources assert the same proposition, produce ONE fact with multiple `source_refs` rather than two near-duplicate facts.

**Rewriting legal-vocabulary facts -- examples:**

| Opinion language (DO NOT use) | Factual rewrite (USE THIS) |
|---|---|
| "Defendant failed to establish the retail-service exemption" | "Defendant produced no payroll records, revenue breakdowns, or customer receipts documenting retail sales to end consumers" |
| "Officer lacked probable cause for the stop" | "Officer initiated the traffic stop without observing a moving violation, equipment defect, or other articulable conduct by the driver" |
| "Plaintiff's regular rate of pay exceeded 1.5x the statutory minimum" | "Plaintiff's weekly commission earnings divided by hours worked exceeded $X.XX in the pay periods examined" |
| "Defendant willfully induced infringement" | "Defendant's internal emails show it received a copy of plaintiff's patent and continued manufacturing the accused product" |
| "Defendants acted pursuant to a pattern and practice of discrimination" | "Defendants stopped Black drivers in the Antelope Valley at 3x the rate of white drivers in the same corridor between 2018 and 2021, per department records" |
| "ECC and ICA qualify as retail or service establishments" | "ECC and ICA derived more than 75% of revenue from sales directly to individual consumers, per their annual sales reports" |

The test: could a layperson investigator verify this fact by reading a document or watching a video, without needing to know any law? If yes, it's a good ground fact.

For each fact, also infer `preferred_embedding`:
- `direct` -- the fact is best demonstrated by a single document that explicitly asserts it (e.g., "Contract X dated March 1, 2019 contains Section 7(b)").
- `pattern` -- the fact emerges only by synthesizing across multiple documents and would be most natural to plant as a multi-doc shard pattern (e.g., "D met with Z three times in March 2019" -- naturally inferred from a calendar entry + an expense report + an email).
- `either` -- works either way.

INPUTS:

OPINION (factual sections only):
{{ opinion_text }}

PLAINTIFF'S MSJ MEMO (if available):
{{ plaintiff_brief_text or "(none)" }}

DEFENSE'S MSJ MEMO (if available):
{{ defense_brief_text or "(none)" }}

PLAINTIFF SOF (if available):
{{ plaintiff_sof_text or "(none)" }}

DEFENSE SOF (if available):
{{ defense_sof_text or "(none)" }}

OPPOSITION BRIEFS / RSOFs (if available):
{{ opposition_text or "(none)" }}

OUTPUT SCHEMA -- JSON array, no markdown, no prose:
[
  {
    "id": "f001",
    "proposition": "<one-sentence factual claim>",
    "source_refs": [
      {"doc_id": "doc045_0", "page": 3, "polarity": "asserts"},
      {"doc_id": "doc053_2", "paragraph": 12, "polarity": "asserts"}
    ],
    "preferred_embedding": "direct|pattern|either",
    "type": "ground"
  }
]

`polarity` is `asserts` if the source affirms the proposition, `refutes` if it contradicts. `type` is always `ground` for this pass; `injected_supporting` and `injected_refuting` are produced by the alt-generation pass.

Aim for **30-80 ground facts** for a typical case. Err on the side of more atomic facts rather than fewer compound ones.
\end{lstlisting}
\end{simplegray}

\paragraph{Plaintiff Hypothesis DAG.}
Loaded by \texttt{extract\_utils/hypothesis\_packet.py}. Constructs the plaintiff's factual narrative as a DAG of sub-claims grouped by Point heading from the plaintiff's brief, where leaves are fact IDs from the canonical fact table.

\begin{simplegray}{Plaintiff-DAG Prompt (\texttt{prompts/extract/plaintiff\_dag.j2})}
\begin{lstlisting}[escapechar=!]
You are extracting the **plaintiff's hypothesis fact-DAG** for a summary judgment case.

The DAG represents the plaintiff's narrative of what happened -- the factual story they tell to support their legal position. The structure is:
- A top-level **claim** (one-sentence summary of the plaintiff's factual narrative)
- A list of **sub-claim groups** that mirror how the plaintiff's brief is organized (typically one group per Point heading in the Memorandum of Law)
- Each group references **fact IDs** from the canonical `facts[]` table -- facts that, taken together, support that sub-claim

You must use ONLY fact IDs that appear in the provided FACT TABLE. Do not invent new facts here; if a fact you need isn't in the table, omit it (a separate pass extracts ground facts).

If the plaintiff's brief lacks clean Point headers (or the brief itself is missing), produce a **flat-list fallback** -- a single sub-claim group with `sub_claim: null` and all relevant fact IDs as `supporting_fact_ids`. This is a documented degraded mode; downstream stages handle it.

INPUTS:

PLAINTIFF'S MSJ MEMO / OPPOSITION BRIEF (the source of the argument structure):
{{ plaintiff_brief_text or "(none -- produce flat-list fallback)" }}

OPINION (for cross-reference; the plaintiff's positions as the judge characterized them):
{{ opinion_text }}

CANONICAL FACT TABLE (use ONLY these fact IDs):
{{ facts_json }}

OUTPUT SCHEMA -- single JSON object, no markdown, no prose:
{
  "claim": "<one-sentence summary of the plaintiff's factual narrative>",
  "fact_dag": [
    {
      "sub_claim": "<one-sentence sub-claim, mirroring a Point heading from the brief>",
      "supporting_fact_ids": ["f001", "f005", "f012"],
      "source_section": "POINT I (or equivalent label from the brief)"
    }
  ],
  "extracted_from": [
    {"doc_id": "doc053_1", "section": "POINT I"},
    {"doc_id": "doc053_1", "section": "POINT II"}
  ]
}

Aim for **2-5 sub-claim groups** typically. If the brief is long and has nested A/B/C sub-headings under each Point, use the Point-level (top-level) groupings, not the sub-headings -- the Point level is what makes a clean fact-DAG node.

If you produce a flat-list fallback, the JSON looks like:
{
  "claim": "...",
  "fact_dag": [
    {"sub_claim": null, "supporting_fact_ids": ["f001", "f002", "..."], "source_section": null}
  ],
  "extracted_from": []
}

DO NOT include facts that aren't in the canonical fact table. DO NOT extract legal conclusions as sub-claims (e.g., "D is liable under Section 271(b)"); sub-claims must be factual ("D had knowledge of the patent before 2018").

**Critical: the `claim` and `sub_claim` fields must be free of legal vocabulary.** A claim should read like a factual summary a journalist would write, not a legal brief. Strip out statute names, legal standards, and conclusory terms. Rewrite to the observable conduct.

**Few-shot examples for `claim`:**

BAD (legal framing): "Plaintiffs were not exempt from FLSA minimum wage and overtime requirements because Defendants failed to establish that ECC and ICA qualify as retail or service establishments and failed to produce evidence that Plaintiffs' regular rate of pay exceeded 1.5 times the statutory minimum wage."
GOOD: "Defendants underpaid plaintiffs -- commission-paid designers and sales reps -- by failing to track hours worked or ensure weekly earnings covered all hours at minimum wage, and never produced records showing sales were primarily retail to end consumers."

BAD (legal framing): "Kelly stopped Plaintiff without probable cause and used excessive force, acting pursuant to the County's longstanding pattern and practice of racially discriminatory policing."
GOOD: "Officer Kelly pulled over Plaintiff without observing a traffic violation, then struck him three times with a baton while he was compliant, in a jurisdiction where department data shows Black drivers were stopped at triple the rate of white drivers on the same roads."

BAD (legal framing): "Defendant willfully induced infringement of the '347 patent by continuing production after receiving plaintiff's cease-and-desist letter."
GOOD: "Defendant continued manufacturing the accused device for 14 months after receiving plaintiff's letter identifying it as a copy of their patented design, while internal emails show engineers discussed the letter and chose not to redesign."
\end{lstlisting}
\end{simplegray}

\paragraph{Defense Hypothesis DAG.}
Loaded by \texttt{extract\_utils/hypothesis\_packet.py}. Mirror of the plaintiff prompt for the side opposing the movant; same DAG structure, same fact-table-only constraint.

\begin{simplegray}{Defense-DAG Prompt (\texttt{prompts/extract/defense\_dag.j2})}
\begin{lstlisting}[escapechar=!]
You are extracting the **defense's hypothesis fact-DAG** for a summary judgment case.

The DAG represents the defendant's narrative of what happened -- the factual story they tell to support their legal position. Structure mirrors the plaintiff DAG: top-level claim, list of sub-claim groups (one per Point heading from the defense brief), each group referencing fact IDs.

You must use ONLY fact IDs that appear in the provided FACT TABLE. Do not invent new facts.

If the defense brief lacks clean Point headers or is missing entirely (e.g., the defense didn't move and the opinion's defense-side argument is only visible as rebuttal paragraphs), produce a **flat-list fallback** -- single sub-claim group with `sub_claim: null` and all relevant fact IDs.

INPUTS:

DEFENSE MSJ MEMO / OPPOSITION BRIEF (the source of the argument structure):
{{ defense_brief_text or "(none -- produce flat-list fallback)" }}

OPINION (for cross-reference; the defense's positions as the judge characterized them):
{{ opinion_text }}

CANONICAL FACT TABLE (use ONLY these fact IDs):
{{ facts_json }}

OUTPUT SCHEMA -- single JSON object, no markdown, no prose:
{
  "claim": "<one-sentence summary of the defense's factual narrative>",
  "fact_dag": [
    {
      "sub_claim": "<one-sentence sub-claim, mirroring a Point heading from the brief>",
      "supporting_fact_ids": ["f003", "f007", "f015"],
      "source_section": "POINT I (or equivalent label)"
    }
  ],
  "extracted_from": [
    {"doc_id": "doc090_0", "section": "POINT I"}
  ]
}

If you must fall back, format:
{
  "claim": "...",
  "fact_dag": [
    {"sub_claim": null, "supporting_fact_ids": ["f003", "f007"], "source_section": null}
  ],
  "extracted_from": []
}

Aim for **2-5 sub-claim groups** typically. Sub-claims must be factual, not legal conclusions.

**Critical: the `claim` and `sub_claim` fields must be free of legal vocabulary.** Strip out statute names, legal standards, and conclusory terms. Rewrite to observable conduct and documents.

**Few-shot examples for `claim`:**

BAD (legal framing): "Defendant established the commissioned-salesperson exemption because plaintiffs' regular rate exceeded 1.5x minimum wage and ECC qualifies as a retail establishment."
GOOD: "Defendants' commission records show plaintiffs earned well above minimum wage in every pay period reviewed, and ECC's sales logs show the majority of transactions were direct retail sales to individual customers."

BAD (legal framing): "Officer Kelly had probable cause for the stop and applied reasonable force consistent with department policy."
GOOD: "Officer Kelly observed Plaintiff's vehicle cross the fog line twice before initiating the stop, and applied a single arm-bar control hold after Plaintiff pulled away from a pat-down, consistent with the force-level documented in his body-camera footage."

BAD (legal framing): "Defendant did not induce infringement because it lacked knowledge of the asserted patent."
GOOD: "Defendant's purchasing records show the accused component was sourced from a third-party supplier whose design predates plaintiff's patent filing, and no internal communications mention plaintiff's patent prior to this lawsuit."

**Posture awareness:** for cases where the defense isn't a movant (e.g., plaintiff-only MSJ, or declaratory-judgment with insurer-as-plaintiff), the "defense" role here means **the side opposing the movant**. The brief used as input is whichever side opposes the MSJ.
\end{lstlisting}
\end{simplegray}

\paragraph{Judge Ruling Annotation.}
Loaded by \texttt{extract\_utils/hypothesis\_packet.py}. Single-pass opinion-only read that tags the already-extracted party hypothesis DAGs with the judge's per-sub-claim ruling and overall disposition.

\begin{simplegray}{Judge-Ruling Prompt (\texttt{prompts/extract/judge\_ruling.j2})}
\begin{lstlisting}[escapechar=!]
You are tagging the judge's ruling onto an already-extracted hypothesis packet. This is a **single-pass, opinion-only** read: we don't re-extract facts or DAGs here, we just annotate which side the judge accepted on each sub-claim, plus the overall disposition.

INPUTS:

JUDGE'S OPINION:
{{ opinion_text }}

PLAINTIFF HYPOTHESIS:
{{ plaintiff_hypothesis_json }}

DEFENSE HYPOTHESIS:
{{ defense_hypothesis_json }}

CANONICAL FACT TABLE (for reference -- fact propositions are here):
{{ facts_json }}

YOUR TASK:

1. Determine the overall disposition:
   - `granted` -- judge granted the movant's MSJ in full
   - `denied` -- judge denied it in full
   - `granted_in_part` -- split ruling
   Identify which side was the **movant** from the opinion's posture section. The movant is the party whose MSJ is being decided.

2. For each sub-claim in the plaintiff's `fact_dag` and the defense's `fact_dag`, decide whether the judge:
   - `accepted` -- judge agreed with this sub-claim's factual narrative
   - `rejected` -- judge disagreed
   - `not_reached` -- judge didn't decide (common when a threshold issue moots later issues, e.g., Pflug where the Release issue mooted 7 of 8 discrimination claims)

3. The side whose **overall claim** the judge accepted gets `judge_ruling: "accepted"` at the hypothesis level. The other side gets `"rejected"`. If a split ruling, mark both as `"split"`.

OUTPUT SCHEMA -- single JSON object, no markdown, no prose:
{
  "posture": {
    "movant_role": "plaintiff|defendant|both",
    "non_movant_role": "plaintiff|defendant|n/a",
    "posture_type": "single_msj|cross_msj|msj_plus_mtd",
    "judge_ruling_movant": "granted|denied|granted_in_part",
    "issues_decided": ["<short label per issue the judge actually reached>"]
  },
  "plaintiff": {
    "judge_ruling": "accepted|rejected|split",
    "sub_claim_rulings": [
      {"sub_claim_index": 0, "ruling": "accepted|rejected|not_reached", "rationale_quote": "<<=1 sentence quote from opinion if obvious>"}
    ]
  },
  "defense": {
    "judge_ruling": "accepted|rejected|split",
    "sub_claim_rulings": [
      {"sub_claim_index": 0, "ruling": "accepted|rejected|not_reached", "rationale_quote": "..."}
    ]
  }
}

The `sub_claim_index` matches the position of the sub-claim in the input `fact_dag` array (0-indexed).

If the opinion is a form order or oral-findings-incorporated order with no written reasoning, return `judge_ruling: "rejected"` (or "accepted") based ONLY on the disposition signal in the docket text and mark every `sub_claim_rulings` entry with `ruling: "not_reached", rationale_quote: "(opinion contains no written reasoning)"`. The case will likely fail the Phase 3 quality filter, and that's the correct outcome.
\end{lstlisting}
\end{simplegray}

\subsection{Corpus Augmentation}
\label{app:prompts:generate}

The prompts in this section take the canonical fact table, the plaintiff/defense hypotheses, and the foundational story produced upstream, and emit alternative factual hypotheses plus the synthetic exhibits that anchor each hypothesis in the augmented corpus.

\paragraph{Alternative Fact DAGs.}
Loaded by \path{generate_utils/alt_fact_dag.py}. Requests two plausible-but-not-actually-true alternative hypotheses per case, each refutable by either an existing ground fact or a planned-to-be-injected refuting fact. 

\begin{simplegray}{Alternative-Hypotheses Prompt (\texttt{prompts/generate/alt\_fact\_dag.j2})}
\begin{lstlisting}[escapechar=!]
You are generating **alternative factual hypotheses** for a summary judgment case. Each alternative is a plausible-but-not-actually-true factual story that:

1. **Re-uses ground facts** -- references >=3 fact IDs from the canonical fact table.
2. **Adds new injected facts** -- does NOT reinterpret existing facts. Instead, introduces NEW facts that, combined with some ground facts, would make the alt look plausible.
3. **Is refutable** -- there is at least one fact (either an existing ground fact, or a new injected refuting fact) that defeats the alt. The model running the benchmark should be able to find this refuter and dissipate the alt.
4. **Coexists with the truth** -- the alt's *supporting* facts are additive: they don't contradict any ground fact. The world stays consistent: ground facts U alt-supporting facts U alt-refuting facts is internally consistent (the only contradiction is the intentional one between alt-supporting and alt-refuting, which is what makes the alt refutable).
5. **Is factual, not legal** -- alts are factual claims about events/states, not legal characterizations or normative judgments.

Generate **2 alts** per case.

INPUTS:

CASE FACT TABLE:
{{ facts_json }}

PLAINTIFF HYPOTHESIS:
{{ plaintiff_hypothesis_json }}

DEFENSE HYPOTHESIS:
{{ defense_hypothesis_json }}

JUDGE_RULING (for context -- the alt should NOT be the judge-rejected hypothesis; it's a third path):
{{ judge_ruling_json }}

ENTITIES:
{{ entities_json }}

FOUNDATIONAL STORY (for realism constraints):
{{ story_excerpt }}

OUTPUT SCHEMA -- JSON array of exactly 2 objects, no markdown, no prose:
[
  {
    "id": "alt_1",
    "claim": "<one-sentence factual hypothesis>",
    "fact_dag": [
      {
        "sub_claim": "<one-sentence sub-claim>",
        "supporting_fact_ids": ["f001", "f005", "<new injected fact ids>"]
      }
    ],
    "ground_facts_referenced": ["f001", "f005"],
    "injected_supporting_facts": [
      {
        "id": "f_alt1_inj_001",
        "proposition": "<NEW factual claim that, if true, would make the alt look plausible>",
        "preferred_embedding": "direct|pattern|either",
        "doc_type_hint": "<email|memo|contract|...>",
        "entity_hint": "<entity that would naturally be the source of this doc>"
      }
    ],
    "injected_refuting_facts": [
      {
        "id": "f_alt1_ref_001",
        "proposition": "<NEW factual claim that defeats the alt>",
        "preferred_embedding": "direct|pattern|either",
        "doc_type_hint": "...",
        "entity_hint": "..."
      }
    ],
    "ground_refuting_fact_ids": [],
    "why_plausible": "<2-3 sentences: why a model on a partial read of the corpus would find this alt tempting>",
    "why_refuted": "<2-3 sentences: what evidence in the corpus dissipates this alt; cite injected_refuting_facts and/or ground_refuting_fact_ids>"
  }
]

**Refutation contract:** every alt must have at least one refuter. Either:
- `injected_refuting_facts` is non-empty (we'll plant new docs in the corpus that refute the alt), OR
- `ground_refuting_fact_ids` is non-empty (an existing ground fact already refutes the alt; no need to inject)
- Both are fine; at least one must be non-empty.

**Additivity check (you must self-verify before outputting):** For each `injected_supporting_facts` and `injected_refuting_facts` entry, confirm: this proposition does NOT contradict any ground fact in the case. (A separate downstream grader will re-check; getting this right here saves regen cost.)

**Anti-pattern (DO NOT generate):** alts that "relitigate" a fact pattern that's already well-documented in the case. E.g., for an SEC v. crypto-token case, do NOT generate "what if the token wasn't a security because the existing transactions had different economic substance" -- the existing transactions are already characterized by the corpus. Better: invent a **new hypothetical transaction** (e.g., "what if there was a separate parallel sale to an institutional buyer that wasn't disclosed") that lives alongside the documented record.

**Diversity:** the 2 alts should explore different angles -- don't generate two near-duplicates. E.g., one might add a new third party and one might add a new time period.
\end{lstlisting}
\end{simplegray}

\paragraph{Multi-Document Shard Plan.}
Loaded by \texttt{generate\_utils/signal\_pipeline.py}. Designs a 3-5 shard plan for embedding a single fact across multiple synthetic documents, such that the full proposition is recoverable only by synthesizing across the bundle.

\begin{simplegray}{Shard-Plan Prompt (\texttt{prompts/generate/shard\_plan.j2})}
\begin{lstlisting}[escapechar=!]
You are designing a **multi-document shard plan** for embedding a single fact across 3-5 documents using the pattern-mode embedding strategy. No single document will assert the fact on its own; instead, each document carries one shard of evidence (a date, an entity, a partial event description) such that the full fact is recoverable only by synthesizing across the bundle.

This is how lawyers build cases in real life: invoices + emails + phone records combine to establish a fact that no single document directly states.

INPUTS:

FACT TO PATTERN-EMBED:
{{ fact_text }}

CASE CONTEXT:

ENTITIES:
{{ entities_json }}

FOUNDATIONAL STORY:
{{ foundational_story_excerpt }}

PREFERRED CASE doc_type DISTRIBUTION (so the shards look natural in this case's corpus):
{{ doc_type_distribution_json }}

YOUR TASK:

Design 3-5 shard documents such that:
1. **Each shard is information-poor on its own** -- no single shard contains the full proposition.
2. **The conjunction asserts the fact** -- reading all shards together, a reader can derive the proposition.
3. **Shards span varied doc_types** -- mix calendar entries, expense reports, emails, internal memos, etc. Pulling the same fact from a single doc_type is too easy and too suspicious.
4. **Shards are temporally and entity-consistent** -- they describe parts of the same real event/state, not unrelated things.

OUTPUT SCHEMA -- single JSON object, no markdown, no prose:
{
  "fact_id": "{{ fact_id }}",
  "fact_text": "{{ fact_text }}",
  "synthesis_check": "<one sentence: what a reader infers from reading all shards together>",
  "shards": [
    {
      "id": "{{ fact_id }}_s1",
      "shard_text": "<the partial fact this shard carries, e.g. 'D and Z had a calendar meeting on March 3, 2019'>",
      "doc_type": "calendar_entry",
      "entity": "<author/source entity for this shard's doc>",
      "time": "2019-03-03",
      "instructions_for_doc_writer": "<1-2 sentences guiding the per-shard signal_fact_anchored.j2 generation>"
    }
  ]
}

**Anti-pattern (DO NOT generate):** shards that each contain enough information to assert the fact alone. If shard 1 already says "D met with Z three times in March 2019 to negotiate a sublicense," there's no synthesis required and the pattern-embedding contract is violated. Each shard must be a partial.

**Self-check before outputting:** for each shard, ask "if a reader sees ONLY this shard and nothing else, can they infer the full fact?" If yes, that shard is too revealing -- split it further or refactor.
\end{lstlisting}
\end{simplegray}

\paragraph{Fact-Anchored Signal Document.}
Loaded by \texttt{generate\_utils/signal\_pipeline.py}. Renders a single synthetic exhibit (email, memo, contract, deposition excerpt, etc.) that asserts or refutes a target fact in the case world, including the per-doc-type formatting partial below.

\begin{simplegray}{Signal-Generation Prompt (\texttt{prompts/generate/signal\_fact\_anchored.j2})}
\begin{lstlisting}[escapechar=!]
You are generating a single synthetic document for a legal-AI benchmark. The document is **anchored to a specific fact** -- your job is to write a realistic `{{ doc_type }}` from `{{ entity }}` dated approximately `{{ time_window }}` that {{ "asserts" if polarity == "asserts" else "refutes" }} the following proposition:

**FACT TO INJECT:**
"{{ fact_text }}"

The document must:
1. Be a coherent, realistic `{{ doc_type }}` that a `{{ entity }}` would plausibly produce in `{{ time_window }}`.
2. Carry the planted proposition clearly enough that a careful reader can verify the fact, but NOT awkwardly stuff the proposition verbatim into the doc.
3. Mention zero other case-specific facts beyond what's needed for context. The reader should have to do work to extract the fact.
4. Be standalone -- no JSON metadata, no headers like "FACT INJECTED:", no debug output.

{% include '_doctype_styles.j2' %}

**Surrounding context (background only -- do NOT inject these):**

CASE FOUNDATIONAL STORY (for tone, era, parties, setting):
{{ foundational_story_excerpt }}

CASE ENTITIES (canonical names; use these for any references):
{{ entities_json }}

**Specific to this generation:**
- doc_type: `{{ doc_type }}`
- author/source entity: `{{ entity }}`
- approximate date: `{{ time_window }}`
- polarity: {{ polarity }} ({{ "the document supports the proposition" if polarity == "asserts" else "the document contradicts the proposition" }})
- fact_id: `{{ fact_id }}`

**Output:** ONLY the rendered document text. No commentary, no markdown wrapping, no explanation. The text will be rendered directly to PDF.

The post-generation grading pass will read this document alone and ask "does it {{ "assert" if polarity == "asserts" else "refute" }} the proposition?" -- make the answer a clear yes.
\end{lstlisting}
\end{simplegray}

\paragraph{Doc-Type Style Partial.}
Included by \texttt{prompts/generate/signal\_fact\_anchored.j2} (and adjacent generators) to enforce per-doc-type formatting -- email headers, memo blocks, contract section numbering, deposition Q/A, expense-report tabular form, prison-form terseness, etc.

\begin{simplegray}{Doc-Type Style Partial (\texttt{prompts/generate/\_doctype\_styles.j2})}
\begin{lstlisting}[escapechar=!]
{# Shared partial: per-doc-type formatting rules. Included by every signal /
   noise / chaff generation prompt to enforce realism. Addresses validator
   complaints about text overrun, weird pipe-separated sections, prison memos
   too verbose for the institution, etc. #}

**Formatting rules by doc_type -- your output MUST match the conventions for the doc_type you were assigned:**

- **`email`** -- Standard email header (From / To / Cc (optional) / Date / Subject) followed by a body of 1-3 short paragraphs. No JSON dumps, no bullet lists in the body, no pipe-delimited sections. End with the sender's name and (optionally) signature block. Length: 100-400 words typical.

- **`memo` / `internal_memo`** -- Header block: `MEMORANDUM\nTo: ...\nFrom: ...\nDate: ...\nRe: ...`. Narrative body 1-3 paragraphs. Optional signature/initials at end. Length: 200-500 words typical.

- **`prison_HSR_form` / `prison_grievance` / `inmate_request_form`** -- Terse and form-driven. NOT narrative prose. Most fields are short labeled answers ("Date: 03/14/2021", "Reason for Request: ..."). One or two sentences per text field, max. The institution that produces these does NOT write multi-paragraph essays.

- **`contract` / `agreement` / `licensing_agreement`** -- Numbered sections (1., 2., 3.), defined terms in boldface or quotes ("Effective Date" means...), recitals at top (WHEREAS clauses), signature block at bottom. Length: 1-5 pages typical, more for complex deals.

- **`deposition_excerpt` / `deposition_transcript`** -- Q/A format strictly: `Q: <question>\nA: <answer>`. Line numbers down the left margin (1, 2, 3 ...). Page numbers. No narrative.

- **`court_filing`** -- Caption block at top (court name, parties, docket number, document title). Numbered paragraphs in the body (1., 2., 3.). Signature block at bottom (attorney name, bar number, address). Length: as appropriate to the doc type.

- **`expense_report` / `invoice` / `receipt`** -- Tabular with line items (date, description, amount). Totals at the bottom. NO narrative paragraphs.

- **`calendar_entry` / `meeting_invite`** -- Single block with subject, date/time, attendees, location, optional brief description (1-2 sentences max). Not a narrative.

- **`text_message` / `sms` / `chat_log`** -- Threaded short messages with timestamps and sender labels. Each message 1-2 sentences max. Casual register.

- **`business_record` / `internal_log`** -- Tabular or chronological entries. Terse. Not narrative.

- **`expert_report`** -- DO NOT GENERATE. Expert content is excluded from the synthetic corpus per the design.

**General rules across all doc_types:**
- Every page must have a visible page number (e.g., `Page 1 of 3` or just `1`).
- No raw JSON, dictionary syntax, pipe-delimited sections, or other debug output bleeding into the rendered text.
- Output should fit cleanly within standard 8.5x11" margins; no text overrun.
- Length must be plausible for the doc_type (don't write a 3-page email; don't write a 1-paragraph contract).
- Date format must be consistent within a single document (don't mix "03/14/2021" and "March 14, 2021" in the same memo).
\end{lstlisting}
\end{simplegray}

\subsection{Factual-QA Harness}
\label{app:prompts:bench}

The factual-QA evaluation runs an agent against each case's question set and grades the agent's free-form response against per-question rubric items.

\paragraph{Factual-QA System Prompt.}
Loaded by \texttt{bench\_utils/run\_single.py}. Pins the agent to the case exhibit directory and enforces the \texttt{ANSWER:} terminator the grader keys on.

\begin{simplegray}{Factual-QA System Prompt (\texttt{prompts/bench/system\_prompt.j2})}
\begin{lstlisting}[escapechar=!]
Look at the following documents, and only make conclusions on the following documents. Do not look online for any outside information of this case, and only rely on what is given to you.

The documents are located in: {{ exhibits_path }}

Answer the following question: {{ question }}

After your analysis, you MUST end your response with your final answer in exactly this format:

ANSWER: <your complete answer to the question>
\end{lstlisting}
\end{simplegray}

\paragraph{Rubric Judge Prompt.}
Loaded by \texttt{bench\_utils/grader.py}. Asks a Claude Sonnet judge to return per-rubric-item booleans for the agent's response, given the gold answer and the rubric items extracted upstream.

\begin{simplegray}{Rubric-Judge Prompt (\texttt{prompts/bench/judge\_prompt.j2})}
\begin{lstlisting}[escapechar=!]
You are a grading judge. You will evaluate whether a response correctly addresses specific rubric items.

Question: {{ question }}
Ground truth answer: {{ answer }}

Response to evaluate:
{{ response }}

For each rubric item below, determine whether the response satisfies it.

Rubric items:
{% for item in rubric_items %}
{{ loop.index }}. {{ item }}
{% endfor %}

Respond with ONLY a JSON array of booleans, one per rubric item, where true means the item is satisfied and false means it is not. Example for 3 items: [true, false, true]
\end{lstlisting}
\end{simplegray}

\subsection{Investigation Harness}
\label{app:prompts:investigation}

The investigation evaluation is a two-turn protocol: in Turn 1 the agent investigates the corpus with tools enabled and produces a self-contained summary; in Turn 2 it commits a verdict on each shuffled hypothesis with tools disabled.

\paragraph{Turn 1: Investigate.}
Loaded by \texttt{bench\_utils/investigation\_runner.py}. Sets the corpus, instructs the agent to identify the most likely account, alternative scenarios, and inconsistent evidence, and warns that follow-up questions will be tool-less.

\begin{simplegray}{Investigation Turn 1 Prompt (\texttt{prompts/bench\_investigation/investigate.j2})}
\begin{lstlisting}[escapechar=!]
You are presented a corpus where a {{ case_type }} matter has happened.

The documents are located in: {{ exhibits_path }}

Please investigate and come up with a solid understanding of what has happened as well as any other possible scenarios the evidence could support. The goal is to be complete in your understanding of what had happened and the possibilities (with cited documents for each possibility).

Do not look online for any outside information. Only rely on the documents in {{ exhibits_path }}.

When you are done investigating, write a thorough summary of:
1. The most likely account of what happened, with citations to specific documents.
2. Any alternative scenarios that the evidence could also support, with citations.
3. Any evidence that appears inconsistent with the most likely account.

Take as long as you need. You will be asked follow-up questions after this summary; you will not be able to re-read documents at that point, so make your summary self-contained.
\end{lstlisting}
\end{simplegray}

\paragraph{Turn 2: Verdict.}
Loaded by \texttt{bench\_utils/investigation\_runner.py}. Presents the shuffled hypothesis menu (\texttt{H1}, \texttt{H2}, \ldots) and forces a believe / don't-believe / don't-know verdict per hypothesis, drawing only on the Turn-1 summary.

\begin{simplegray}{Investigation Turn 2 Prompt (\texttt{prompts/bench\_investigation/verdict.j2})}
\begin{lstlisting}[escapechar=!]
Now that you've conducted your investigation, please answer the following questions on your beliefs about each hypothesis. For each, choose exactly one of:

  A. Believe
  B. Don't believe
  C. Don't know

The hypotheses:
{% for item in hypotheses %}
Hypothesis {{ item.label }}: {{ item.text }}
{% endfor %}

Rely only on what you learned during your investigation. You cannot search the documents again.

Respond with ONLY a JSON array -- no prose, no code fences -- with one object per hypothesis in the same order as above:

[
  {"label": "H1", "verdict": "A", "reason": "one short sentence"},
  {"label": "H2", "verdict": "B", "reason": "one short sentence"},
  ...
]
\end{lstlisting}
\end{simplegray}


\end{document}